\documentclass[letterpaper]{article} 
\usepackage[preprint]{aaai2027}  
\usepackage[hyphens]{url}  
\usepackage{graphicx} 
\usepackage{natbib}  
\usepackage{caption} 
\usepackage{pifont}
\usepackage{booktabs}
\usepackage{multirow}
\usepackage{array}
\usepackage{xcolor}
\usepackage{graphicx}
\usepackage{makecell}
\usepackage{enumitem}
\usepackage{amssymb}
\usepackage{subcaption}
\usepackage{algorithm}
\usepackage{algorithmic} 

\usepackage{amsmath}
\usepackage{amssymb}
\usepackage[most]{tcolorbox}
\usepackage{tikz}
\usepackage{pdfpages}

\newtcolorbox{modelbox}[1]{
    colback=white,
    colframe=brown!80!black,
    fonttitle=\bfseries,
    title=#1,
    arc=5pt,
    enhanced,
    attach boxed title to top left={yshift=-2mm, xshift=5mm},
    boxed title style={colback=brown!80!black},
    after title=\vspace{1mm}
}
\usepackage{listings}
\usepackage{xcolor}

\usepackage{listings}
\usepackage{xcolor}

\usepackage{newfloat}
\usepackage{listings}
\DeclareCaptionStyle{ruled}{labelfont=normalfont,labelsep=colon,strut=off} 
\floatstyle{ruled}
\newfloat{listing}{tb}{lst}{}
\floatname{listing}{Listing}

\usepackage{booktabs}

\title{Not All Problems Are Best Modeled as MILP: \\ A DSL-Centric Framework for Flexible and Accurate Optimization Modeling}
\author{
  Shaofeng Zhang$^{1,4}$\equalcontrib \quad
  Hongyuan Su$^{2,4}$\equalcontrib \quad
  Qingwen Peng$^{3,4}$ \equalcontrib \quad
  Zefang Zong$^{2}$ \\[0.4em]
  Shengcai Liu$^{1}$ \quad
  Ke Tang$^{1}$ \quad
  Yong Li$^{2}$ \corresponding
}
\affiliations{
  $^{1}$Southern University of Science and Technology, Shenzhen \\
  $^{2}$Tsinghua University, Beijing \\
  $^{3}$Tianjin University, Tianjin \\
  $^{4}$Zhongguancun Academy, Beijing \\
}

\begin{document}

\maketitle

\begin{abstract}
  Solving combinatorial optimization problems (COPs) requires not only efficient algorithms but also carefully crafted formulations. While recent works have leveraged LLMs to automate optimization modeling, current frameworks predominantly rely on a rigid mixed-integer linear programming (MILP) paradigm. In this paper, we argue that not all problems are best modeled as MILP, as forcing complex domains into linear constraints can induce prohibitive modeling complexity and severely restrict solver flexibility. To address this, we propose OptiDSL, a framework that shifts the focus from rigid MILP formulations to domain-specific language (DSL) representations. By utilizing LLMs to map natural language onto standardized, domain-accepted structures, OptiDSL decouples problem formulation from execution. This paradigm enables seamless integration with a diverse library of specialized solvers, ranging from traditional heuristics to modern learning-based methods. Experimental results on the comprehensive benchmark of 44 COP types show that OptiDSL significantly surpasses MILP-based pipelines, yielding a 51.66\% gain in formulation accuracy and a 91.71\% decrease in modeling time. Notably, it also outperforms MILP-based pipelines on the existing benchmark, achieving a 23.09\% higher formulation accuracy. 
  Our code is available at \textcolor{blue}{\url{https://anonymous.4open.science/r/OptiDSL}}.
\end{abstract}


\section{Introduction}
Combinatorial optimization problems (COPs) encompass a broad range of real-world decision-making tasks, from manufacturing scheduling~\citep{wang2021review} to logistics and operations planning~\citep{zong2025DDS}. While solving COPs efficiently has been a long-standing research focus, a more fundamental and often overlooked challenge lies in their formulation: translating high-level requirements into precise mathematical models that solvers can process. Traditionally, this formulation process is carried out manually by domain experts, who must encode the optimization objectives and constraints and then select and execute an appropriate solver. Due to the diverse and domain-specific nature of COPs, this manual pipeline is not only time-consuming but also error-prone, presenting a major barrier to broader adoption and full automation of optimization workflows.

The emergence of large language models (LLMs)~\citep{chowdhery2023palm,brown2020language} presents an opportunity to bridge this gap. By leveraging LLMs, it becomes possible to process problem descriptions in natural language and automatically translate them into formal optimization formulations. Recent efforts have explored this direction by mapping arbitrary problem statements into unified formulations via Linear Programming (LP) or Mixed-integer LP (MILP), and further solved via corresponding MILP solvers~\citep{gurobi2024,cplex2009}.


\begin{figure*}[t]
    \centering
    \includegraphics[width=1.\textwidth]{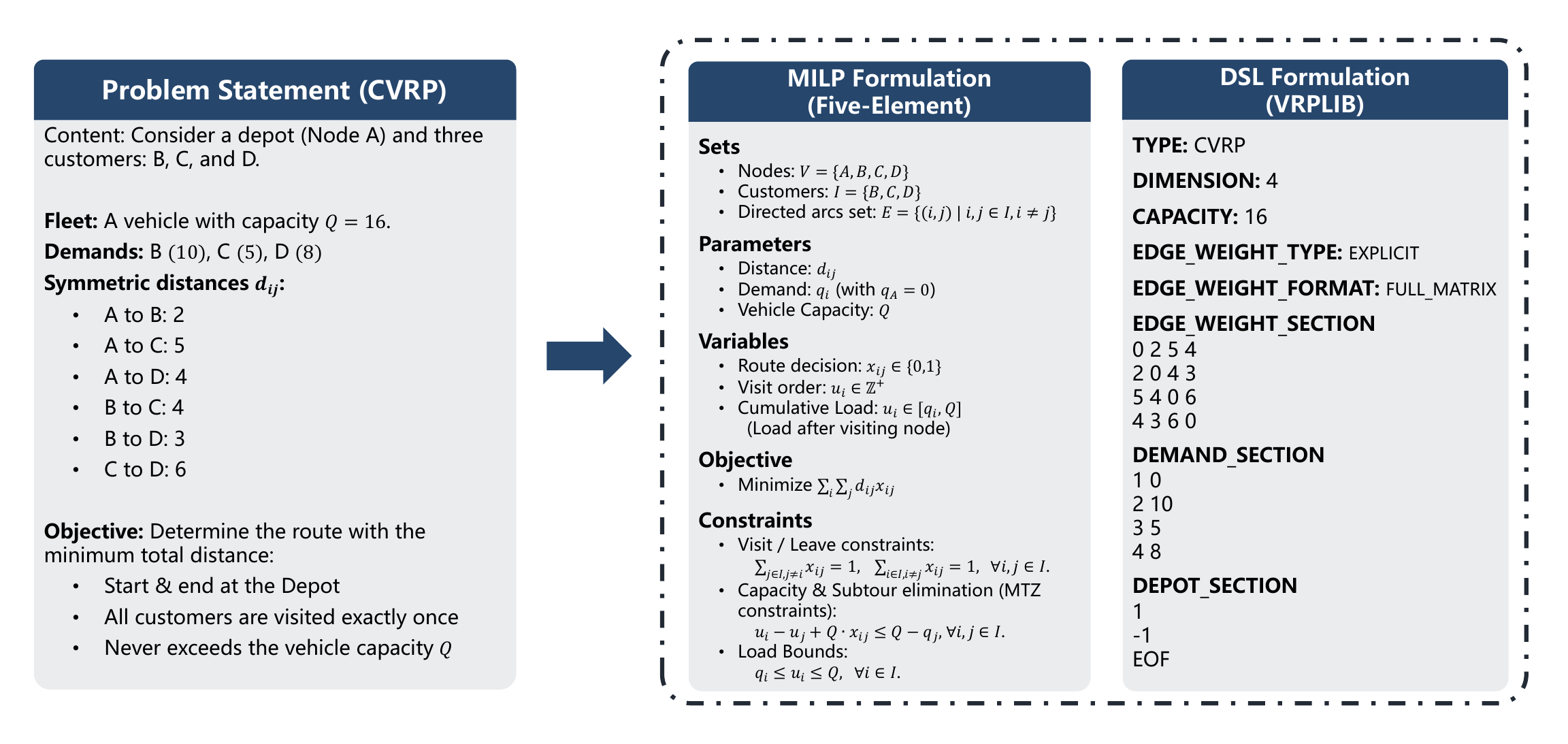}
    \caption{An illustrative example of modeling paradigms for the CVRP case. The DSL formulation is more appropriate than the MILP formulation for representing CVRP instances.}
    \label{fig:example}
\end{figure*}

While the MILP modeling paradigm has shown promise, it faces significant limitations when applied to complex COPs. First, the MILP paradigm struggles with modeling complexity in certain combinatorial domains. 
For instance, when formulating the capacitated vehicle routing problem (CVRP), the Dantzig-Fulkerson-Johnson formulation~\citep{DFJF1960} serves as the representative  modeling scheme to eliminate subtours and ensure solution feasibility. However, this approach requires a number of constraints that grows exponentially with problem scale, as illustrated in Figure~\ref{fig:example}.
This explosion in constraints frequently exceeds the context and reasoning capabilities of LLMs, leading to a sharp decline in modeling accuracy. Second, rigid adherence to MILP restricts the choice of solvers, preventing the utilization of efficient, domain-specific algorithms that offer significantly higher computational efficiency than MILP solvers.


In this paper, we argue that not all optimization problems are best formulated as MILP and present \textbf{OptiDSL}, a framework centered on a domain-specific language (DSL) modeling workflow. Here, the DSL refers to the widely recognized and adopted format tailored for describing problem instances within specialized domains. Moreover, almost all solvers developed for these domains are natively compatible with this DSL template.
For example, in routing problems, it utilizes VRPLib-style~\citep{pyvrp2023} structures to transform natural language into standardized, solver-ready formats. By adopting these templates as the intermediate representation, OptiDSL effectively decouples problem formulation from solver execution and enables automatic matching of problem instances with a diverse library of specialized solvers, ranging from traditional heuristics to modern learning-based methods.



To support systematic evaluation, we introduce a comprehensive benchmark dataset spanning 44 common COP types, each featuring natural language descriptions and structured instance data. This benchmark is constructed in a semi-automated manner by expanding a manually curated set of seed examples into diverse variants using controlled language generation techniques.

Our main contributions are summarized as follows:
\begin{itemize}
\item \textbf{OptiDSL Workflow Development}: 
We propose \textbf{OptiDSL}, a DSL-based framework that overcomes the expressive limitations of MILP by mapping natural language into specialized DSL templates. This framework enables seamless adaptation to diverse specialized solvers where MILP remains inefficient or inapplicable.
\item \textbf{Comprehensive Dataset Construction}: We introduce a comprehensive benchmark covering 44 COP types across domains like VRP, Scheduling, and Packing. By providing diverse natural language descriptions and structured instances, this dataset ensures a realistic and comprehensive assessment of modeling capabilities across varied COPs.
\item \textbf{Extensive Experimental Evaluation}: Our evaluation demonstrates that OptiDSL significantly outperforms MILP-only paradigms. On the comprehensive benchmark, it achieves a 51.66\% gain in formulation accuracy and a 91.71\% reduction in modeling time, exhibiting superior scalability and effectiveness. Notably, it also outperforms MILP-based pipelines on the existing benchmark, achieving a 23.09\% higher formulation accuracy.
\end{itemize}

\section{Related Works}

\begin{figure*}[htbp]
    \centering
    \includegraphics[width=0.8\textwidth]{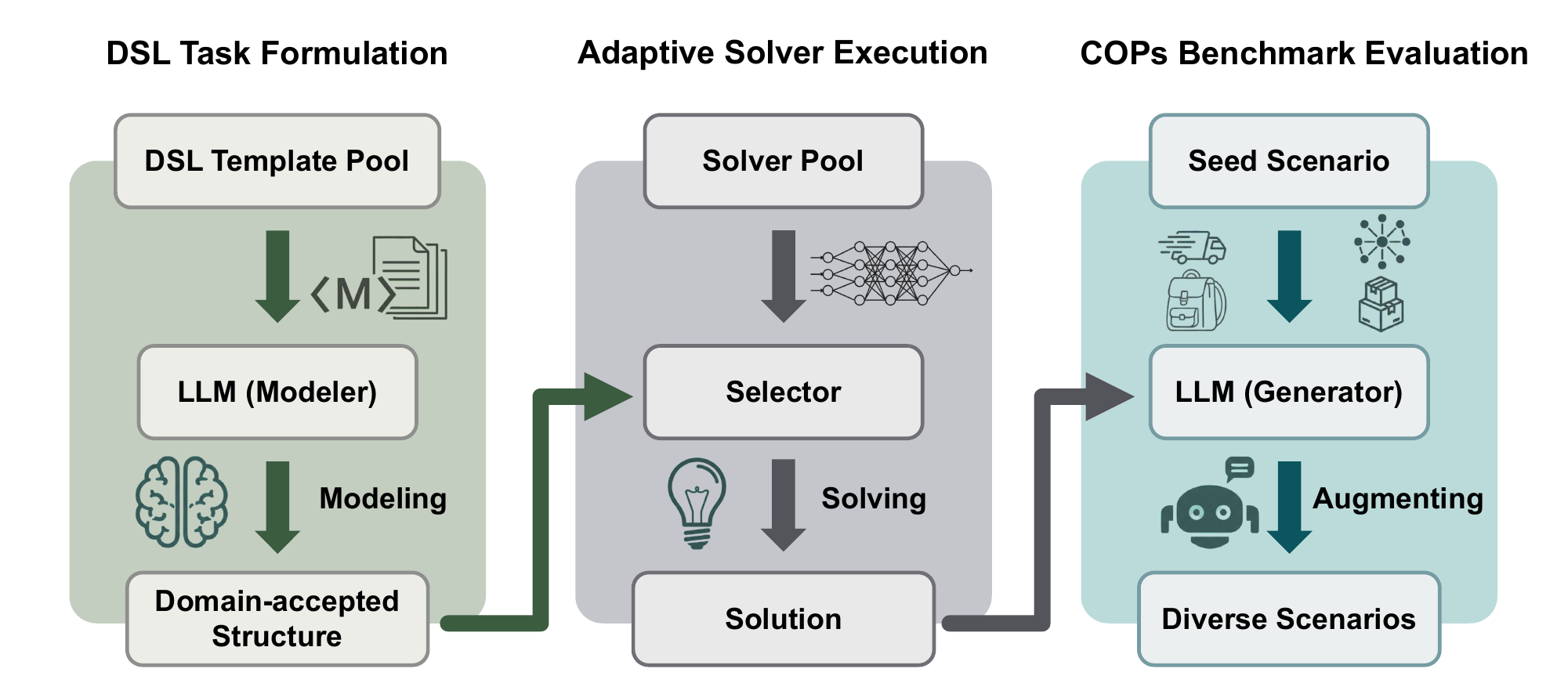}
    \caption{\textbf{OptiDSL Workflow:} The framework begins with DSL Task Formulation (left), where an LLM translates unstructured natural language descriptions into standardized DSL data templates. These structured instances are then passed to the Adaptive Solver Execution module (center), which dynamically routes the tasks to the optimal solvers based on performance trade-offs. Finally, the COPs Benchmark Evaluation (right) systematically assesses the generated formulations and solutions across diverse problem scenarios.}
    \label{fig:workflow}
\end{figure*}

\subsection{Solvers for COPs}

Existing COP solvers generally fall into two categories based on optimality and specificity. First, exact MILP solvers~\citep{gurobi2024, cplex2009} guarantee mathematical optimality but suffer from prohibitive computational costs and poor scalability on large instances~\citep{kool2018attention}. Moreover, reformulating diverse COPs into rigid MILP matrices is notoriously complex~\citep{NL4Opt, CoE}.

Conversely, flexible algorithms trade strict optimality for significant computational efficiency by exploiting domain-specific strategies. This category spans traditional search heuristics~\citep{mu2016Simulated, sohrabi2023genetic, Flerova2015, Consolini2019} and modern Neural Combinatorial Optimization (NCO) methods~\citep{sun2023difusco, berto2024routefinder, Jin2024Unified, bello2017neural, kwon2020pomo}. While these solvers rapidly deliver high-quality solutions, they natively require domain-specific data formats (DSLs) rather than standard MILP matrices, creating a critical integration barrier for generalized automated pipelines.

\subsection{Auto Formulating-Solving COPs Methods}
Recent benchmarks in automated combinatorial optimization focus predominantly on translating natural language descriptions into Mixed-Integer Linear Programming (MILP) formulations. Early work pioneered this direction by decoupling the formulation process into semantic entity recognition and logic generation \citep{NL4Opt}. For lengthy and complex scenarios, subsequent frameworks have leveraged multi-agent cooperative reasoning \citep{CoE, OptiMUS}. To address data scarcity, other methods enhance formulation capabilities via data synthesis, employing either automated data augmentation \citep{ORLM} or solver-verified synthetic datasets \citep{MAMO}. More recently, unified frameworks have utilized multi-instruction tuning and self-correction mechanisms to improve modeling accuracy and generalization across diverse real-world domains \citep{jiang2025llmopt}.

However, the previous text-to-MILP paradigm presents two critical limitations. First, formulating complex combinatorics (e.g., subtour elimination) as MILP constraints creates an expressive bottleneck, generating an exponential number of constraints that overwhelm the reasoning capacities of current LLMs. Second, standard MILP matrices inherently preclude the use of highly efficient flexible solvers (such as search heuristics and neural methods), which require domain-specific inputs. \textbf{OptiDSL} overcomes these limitations via a \textit{text-to-DSL} paradigm. By leveraging domain-specific languages (e.g., VRPLib format), OptiDSL circumvents the mathematical complexity of MILP modeling and naturally unlocks seamless integration with a diverse pool of both exact algorithms and flexible solvers.

\begin{table*}[t]
\centering
\small

\renewcommand{\arraystretch}{0.82}
\setlength{\tabcolsep}{3pt}
\setlength{\extrarowheight}{0pt}

\begin{tabular}{cccc}

\specialrule{0.8pt}{0pt}{0.5pt}
\specialrule{0.8pt}{0pt}{1pt}

\multirow[c]{2}{*}{\textbf{Domain}}
&
\multirow[c]{2}{*}{\textbf{Problem Type}}
&
\multicolumn{2}{c}{\textbf{Solver}}
\\[-1pt]

\cmidrule(lr){3-4}

&
&
\textbf{Non-Learning}
&
\textbf{Learning}
\\

\midrule

\multirow[c]{8}{*}{VRP}
& CVRP, OVRP, VRPB, VRPL
& \multirow[c]{2}{*}{
    \makecell[c]{Gurobi\\[-2pt]\citep{gurobi2024}}
}
& \multirow[c]{8}{*}{
    \makecell[c]{RouteFinder\\[-2pt]\citep{berto2024routefinder}}
}
\\

& VRPTW, OVRPTW, OVRPB
&
&
\\

& OVRPL, VRPBL, VRPBTW
& \multirow[c]{2}{*}{
    \makecell[c]{LKH\\[-2pt]\citep{lkh}}
}
&
\\

& VRPLTW, OVRPBL, OVRPBTW
&
&
\\

& OVRPLTW, VRPBLTW, OVRPBLTW
& \multirow[c]{2}{*}{
    \makecell[c]{PyVRP\\[-2pt]\citep{pyvrp2023}}
}
&
\\

& VRPMB, OVRPMB, VRPMBL
&
&
\\

& VRPMBTW, OVRPMBL, OVRPMBTW
& \multirow[c]{2}{*}{
    \makecell[c]{OR-Tools\\[-2pt]\citep{ortools2019}}
}
&
\\

& VRPMBLTW, OVRPMBLTW
&
&
\\

\midrule

\multirow[c]{5}{*}{SP}
&
\multirow[c]{5}{*}{
    \makecell[c]{
        JSSP, FJSSP,\\[-2pt]
        FSSP, HFSSP,\\[-2pt]
        OSSP, ASP
    }
}
& Gurobi
& \makecell[c]{L2D\\[-2pt]\citep{L2D2020}}
\\

&
&
\makecell[c]{CP-SAT\\[-2pt]\citep{Perron2023cpsat}}
&
\makecell[c]{FJSP-DRL\\[-2pt]\citep{FJSPDRL2023}}
\\

&
&
\makecell[c]{
    Dispatching Rules\\[-2pt]
    \citep{dispatchingRules1982}
}
&
\makecell[c]{DANIEL\\[-2pt]\citep{DANIEL2024}}
\\

&
&
\multirow[c]{2}{*}{
    \makecell[c]{
        Genetic Algorithm (GA)\\[-2pt]
        \citep{Zhang2011spga}
    }
}
&
\makecell[c]{MatNet\\[-2pt]\citep{MatNet2021}}
\\

&
&
&
\makecell[c]{GOAL\\[-2pt]\citep{goal2025}}
\\

\midrule

\multirow[c]{2}{*}{BPP}
& 2DBPP, 2DBPPR
& Gurobi
& \multirow[c]{2}{*}{
    \makecell[c]{PCT\\[-2pt]\citep{zhao2022learning}}
}
\\

& 3DBPP, 3DBPPR
& Genetic Algorithm (GA)
&
\\

\midrule

\multirow[c]{2}{*}{GP}
& MIS, MVC
& \multirow[c]{2}{*}{Gurobi}
& \makecell[c]{FastT2T\\[-2pt]\citep{li2024fast}}
\\

& Max Cut, Max Clique
&
& \makecell[c]{DiffUCO\\[-2pt]\citep{diffuco2024}}
\\

\midrule

\multirow[c]{2}{*}{KP}
& 0-1KP, BKP, UKP
& Gurobi
& \multirow[c]{2}{*}{
    \makecell[c]{POMO\\[-2pt]\citep{kwon2020pomo}}
}
\\

& MD0-1KP, MDBKP, MDUKP
& Dynamic Programming (DP)
&
\\

\specialrule{0.8pt}{1pt}{0.5pt}
\specialrule{0.8pt}{0pt}{0pt}

\end{tabular}

\caption{\textbf{Overview of the OptiDSL Solver Pool:} Covering 5 problem domains and 44 distinct COPs, the framework integrates diverse domain-specific solvers. These are categorized into classical non-learning algorithms and modern learning-based methods.}
\label{tab:solverpool}
\end{table*}

\section{Preliminary}
The core problem addressed in this paper is the automated formulation of COPs from unstructured text. Formally, given a natural language description $\mathcal{T}$ of a COP, our objective is to automate both its formulation and solution processes. Unlike previous approaches that rigidly restrict the pipeline to MILP formulations and exact MILP solvers, our work adopts a \textit{text-to-DSL} modeling paradigm. Different families of COPs inherently possess their own specialized modeling formats. By utilizing DSLs as established domain standards, our formulation naturally adapts to a diverse array of specialized solvers. This decoupling allows us to flexibly alternate between \textit{exact solvers} and \textit{flexible solvers}. While exact solvers guarantee optimal solutions but often incur prohibitive computational costs, flexible solvers (e.g., NCO solvers) prioritize computational efficiency, rapidly delivering high-quality solutions despite lacking strict optimality guarantees.

To evaluate our proposed framework, this work selects five broad and widely studied combinatorial optimization tasks. The specific domains and their corresponding problem types are detailed as follows:
\begin{itemize}[leftmargin=*]
\item \textbf{Capacitated Vehicle Routing Problems (CVRP)}: Optimize vehicle routes to serve customers while minimizing cost and satisfying constraints like capacity and time windows. We consider 24 VRP variants with control variables including Open Route (O), Backhaul (B), Mixed Backhaul (MB), Duration Limit (L), and Time Windows (TW).

\item \textbf{Scheduling Problems (SP)}: Assign jobs to machines over time, respecting operation precedence and machine constraints, with objectives such as minimizing makespan. We include six variants: Job Shop, Flexible Job Shop, Flow Shop, Hybrid Flow Shop, Open Shop, and Assembly Scheduling.

\item \textbf{Bin Packing Problems (BPP)}: Pack items into bins with fixed capacity while minimizing bin count. We examine eight variants derived from combinations of Dimension (2D/3D) and Orientation Constraint (rotatable/non-rotatable).

\item \textbf{Graph Problems (GP)}: Optimize structural objectives in graphs, such as cuts or independent sets. We focus on four problems: Maximum Independent Set, Minimum Vertex Cover, Max Cut, and Max Clique.

\item \textbf{Knapsack Problems (KP)}: Select a subset of items with given weights and values to maximize total value without exceeding the knapsack's capacity.
\end{itemize}

\section{OptiDSL}

To automate and efficiently solve COPs, we introduce \textbf{OptiDSL}, a framework that leverages DSLs to decouple problem formulation from downstream solvers, thereby overcoming the constraint explosion of rigid MILP paradigms. As depicted in Figure~\ref{fig:workflow}, OptiDSL comprises three core components: (1) DSL-Based Task Formulation, translating natural language into structured DSL templates; (2) Adaptive Solver Execution, dynamically routing instances to specialized algorithms; and (3) COPs Benchmark Evaluation, systematically assessing performance via our proposed \textit{OptiDSLBench}.

\subsection{DSL-Based Task Formulation} \label{sec:formulation} 

Instead of mapping natural language to the complex MILP format, OptiDSL leverages pre-defined DSL templates. Representing widely adopted paradigms in optimization, these DSLs are highly standardized, as shown in Figure~\ref{fig:example}. Our candidate pool $\mathcal{T} = \{T_1, \dots, T_k\}$ spans five major COP domains derived from established benchmarks: VRP from VRPLIB~\citep{pyvrp2023}; BPP~\citep{HOPPER200134}, SP~\citep{TAILLARD1993278}, and KP~\citep{ChuBeasley1998} from OR-Library; and GP from Network Repository~\citep{nr-aaai15}. Adopting these domain standards as our intermediate representation circumvents the constraint explosion plaguing MILP formulations and significantly reduces the LLM's reasoning burden.

To map a problem description $m$ into a structured DSL file $d$, we design a hierarchical agentic workflow for the LLM $\mathcal{L}$. The process begins with DSL semantic routing to mitigate context bloat. Instead of loading all structural rules simultaneously, each candidate DSL in $\mathcal{T}$ is represented by a concise meta-description. The LLM acts as a routing agent, aligning the semantic profile of $m$ with these descriptions to autonomously dispatch the task to the appropriate template $T^*$. Following this selection, the workflow advances to the DSL instantiation phase. Governed by the rigorous syntax of $T^*$, the agent structures the natural language text into a formalized problem instance. Beyond naive data extraction, the agent performs logical deduction to ground natural language into symbolic constraints. For example, in the open vehicle routing scenario, the problem description may simply state that ``vehicles are not required to return to the starting point'' without explicitly providing a boolean value. Consequently, the model deduces this return-to-depot requirement and subsequently toggles the `OPEN\_ROUTE' flag in the DSL format file.



This two-stage formulation ensures precise constraint capture while minimizing LLM context overhead. Furthermore, the COPs curated in our framework serve as fundamental problem classes. Therefore, the corresponding DSLs provide foundational formats upon which more complex real-world optimization problems can be seamlessly extended (e.g., imposing time windows on the standard CVRP to formulate CVRPTW). By leveraging the reasoning capabilities of the LLM, OptiDSL dynamically adapts these base templates to accommodate novel constraints. This endows the framework with strong generalization, ensuring its broad applicability and universality across diverse practical optimization domains.

\subsection{Adaptive Solver Execution} \label{sec:execution}

Unlike the rigid MILP formulation paradigm, OptiDSL standardizes the intermediate representation via DSLs to effectively decouple problem formulation from downstream solvers. This architectural flexibility enables seamless integration with a versatile solver pool $\mathcal{S}$ (see Table~\ref{tab:solverpool}).

Since exhaustive parallel execution of all solvers incurs prohibitive computational overhead, we introduce an adaptive solver routing mechanism driven by offline empirical profiling. For each DSL domain, we benchmark all compatible solvers $s \in \mathcal{S}$ on a set of representative instances. This establishes multi-dimensional performance profiles capturing metrics like optimality gap and computational time. As demonstrated in Table~\ref{tab:solver_cvrp}, solvers inherently exhibit distinct trade-offs between solving time and solution quality. For example, exact MILP solvers provide strict theoretical guarantees for optimality but incur exponentially high time costs, whereas NCO solvers offer exceptional computational efficiency but lack theoretical bounds. Consequently, different problem scales and deployment constraints necessitate distinct solver selection strategies. By capturing these varying dynamics, the empirical profiles effectively map the Pareto-optimal front for each problem domain.

During online execution, the framework leverages these profiles to dynamically route the DSL data $d$ to the optimal solver $s^*$, guided by user-specified preferences $P$. Whether prioritizing rapid execution for real-time dispatching or maximum solution quality for strategic planning, OptiDSL adaptively selects the best solver. This preference-aware execution ensures a Pareto-efficient balance between computational cost and performance, unlocking efficiency unattainable by monolithic MILP pipelines.

\subsection{COPs Benchmark Evaluation}

To systematically evaluate our framework, we introduce \textbf{OptiDSLBench}, a comprehensive benchmark spanning 44 common COP types. To construct diverse, high-fidelity natural language descriptions that reflect real-world requirements, we employ a three-stage semi-automated generation pipeline:

\begin{itemize}[leftmargin=*]
    \item \textbf{Generate}: For each COP type, we initialize a seed dataset of scenarios. We then prompt an LLM to expand this set with novel descriptions, explicitly restricting generated titles from duplicating existing ones to ensure semantic variety.

    \item \textbf{Check \& Modify}: To mitigate LLM hallucinations~\citep{LLMHallucination2025} that may omit critical constraints, we deploy a scenario-checking agent. It verifies whether a generated scenario logically aligns with its formal COP definition. Invalid scenarios are manually revised based on the agent's rationale, while valid ones undergo random sampling audits to guarantee dataset quality.

    \item \textbf{Data Placeholder Substitution}: Because large-scale numerical data generation via LLMs is error-prone, we decouple structural generation from data instantiation. The placeholder tags (e.g., $\langle \text{demand} \rangle$) are explicitly defined by the user. Guided by the problem's semantic context and the provided DSL field descriptions, the LLM automatically identifies and matches these user-defined placeholders during formulation, focusing purely on constraint validity.
\end{itemize}

\begin{table*}[htbp]
    \centering
    \scriptsize
    \renewcommand{\arraystretch}{0.9}
    \setlength{\tabcolsep}{5pt}
    \small
    \renewcommand{\arraystretch}{0.5}
    \begin{tabular}{ll ccc ccc ccc ccc}
        \toprule
        \multirow{2}{*}{Domain} & \multirow{2}{*}{Type} 
        & \multicolumn{3}{c}{Our Method} 
        & \multicolumn{3}{c}{COE} 
        & \multicolumn{3}{c}{ORLM} 
        & \multicolumn{3}{c}{LLMOPT} \\
        \cmidrule(lr){3-5} \cmidrule(lr){6-8} \cmidrule(lr){9-11} \cmidrule(lr){12-14}
        & & ER$\uparrow$ & OR$\uparrow$ & MT$\downarrow$
        & ER$\uparrow$ & OR$\uparrow$ & MT$\downarrow$
        & ER$\uparrow$ & OR$\uparrow$ & MT$\downarrow$
        & ER$\uparrow$ & OR$\uparrow$ & MT$\downarrow$ \\ 
        \midrule

        \multirow{6}{*}{\textbf{VRP}}
        & CVRP      & \textbf{0.98} & \textbf{0.93} & \textbf{10.61} & 0.94 & 0.46 & 147.53 & 0.77 & 0.28 & 81.86 & 0.30 & 0.00 & 120.35 \\
        & OVRP      & \textbf{0.98} & \textbf{0.94} & \textbf{10.66} & 0.93 & 0.17 & 152.00 & 0.79 & 0.09 & 84.80 & 0.30 & 0.02 & 116.95 \\
        & OVRPB     & \textbf{0.91} & \textbf{0.60} & \textbf{11.47} & 0.86 & 0.04 & 144.20 & 0.76 & 0.03 & 90.33 & 0.28 & 0.00 & 128.40 \\
        & OVRPBL    & \textbf{0.87} & \textbf{0.51} & \textbf{11.70} & 0.85 & 0.02 & 240.53 & 0.75 & 0.01 & 155.02 & 0.22 & 0.00 & 151.96 \\
        & VRPMBL    & \textbf{0.86} & \textbf{0.68} & \textbf{11.52} & 0.81 & 0.18 & 166.08 & 0.70 & 0.14 & 94.62 & 0.18 & 0.00 & 130.23 \\

        & \multicolumn{10}{c}{\dots} \\
        & \multicolumn{10}{c}{(see complete results for all 24 types of VRP in supplementary material)}\\
      
        \midrule
        \multirow{6}{*}{\textbf{SP}}
        & ASP       & \textbf{1.00} & \textbf{0.84} & \textbf{5.92}             & 0.92 & 0.47 & 75.64 & 0.84 & 0.45 & 42.93 & 0.29 & 0.17 & 88.49 \\
        & JSSP      & \textbf{1.00} & \textbf{0.96} & \textbf{6.38} & 0.93 & 0.54 & 67.85 & 0.91 & 0.40 & 37.61 & 0.33 & 0.09 & 81.40 \\
        & FJSSP     & \textbf{1.00} & \textbf{0.88} & \textbf{8.68} & 0.91 & 0.53 & 71.42 & 0.86 & 0.48 & 38.77 & 0.24 & 0.09 & 94.57 \\
        & FSSP      & \textbf{1.00} & \textbf{0.74} & \textbf{6.12} & 0.90 & 0.65 & 66.40 & 0.85 & 0.62 & 34.16 & 0.44 & 0.20 & 78.35 \\
        & OSSP      & \textbf{1.00} & \textbf{1.00} & \textbf{14.78} & 0.97 & 0.56 & 64.09 & 0.92 & 0.43 & 42.14 & 0.45 & 0.13 & 76.70 \\
        & HFSSP     & \textbf{0.97} & \textbf{0.74} & \textbf{14.56} & 0.93 & 0.29 & 69.38 & 0.85 & 0.25 & 39.08 & 0.22 & 0.01 & 95.52 \\
        \midrule
        
        \multirow{4}{*}{\textbf{BPP}}
        & 2DBPP   & \textbf{1.00} & \textbf{0.99} & \textbf{4.35} & 0.88 & 0.41 & 86.87 & 0.67 & 0.25 & 52.52 & 0.45 & 0.26 & 94.19 \\
        & 2DBPPR  & \textbf{1.00} & \textbf{0.99} & \textbf{3.86} & 0.88 & 0.44 & 84.93 & 0.57 & 0.23 & 52.51 & 0.40 & 0.28 & 87.80 \\
        & 3DBPP   & \textbf{1.00} & \textbf{0.85} & \textbf{4.51} & 0.83 & 0.44 & 95.99 & 0.54 & 0.21 & 66.75 & 0.62 & 0.39 & 77.51 \\
        & 3DBPPR  & \textbf{1.00} & \textbf{1.00} & \textbf{4.44} & 0.88 & 0.48 & 92.14 & 0.63 & 0.25 & 54.45 & 0.55 & 0.37 & 84.21 \\
        \midrule
        
        \multirow{6}{*}{\textbf{KP}}
        & 0-1KP     & \textbf{1.00} & \textbf{0.95} & \textbf{4.17} & 0.95 & 0.70 & 61.19 & 0.91 & 0.73 & 32.77 & 0.97 & 0.81 & 55.05 \\
        & BKP       & \textbf{1.00} & \textbf{0.87} & \textbf{4.45} & 0.96 & 0.68 & 67.62 & 0.92 & 0.53 & 38.01 & 0.98 & 0.76 & 64.67 \\
        & UKP       & \textbf{1.00} & \textbf{0.83} & \textbf{3.84} & 0.92 & 0.75 & 61.85 & 0.89 & 0.63 & 31.52 & \textbf{1.00} & 0.82 & 49.04 \\
        & MD0-1KP   & \textbf{1.00} & \textbf{0.97} & \textbf{4.56} & \textbf{1.00} & 0.78 & 63.55 & 0.95 & 0.91 & 39.40 & 0.97 & 0.93 & 59.65 \\
        & MDBKP     & \textbf{1.00} & \textbf{0.98} & \textbf{4.55} & 0.95 & 0.81 & 69.88 & 0.89 & 0.69 & 44.06 & 0.92 & 0.69 & 75.69 \\
        & MDUKP     & \textbf{1.00} & \textbf{0.94} & \textbf{4.49} & 0.91 & 0.83 & 63.57 & 0.86 & 0.63 & 32.01 & \textbf{1.00} & 0.86 & 63.53 \\
        \midrule
        
        \multirow{4}{*}{\textbf{GP}}
        & MVC       & \textbf{1.00} & \textbf{1.00} & \textbf{3.85} & \textbf{1.00} & 0.90 & 47.05 & 0.78 & 0.76 & 31.16 & 0.94 & 0.82 & 59.51 \\
        & MIS       & \textbf{0.99} & \textbf{0.99} & \textbf{3.93} & 0.98 & 0.88 & 44.96 & 0.82 & 0.68 & 23.88 & 0.91 & 0.77 & 62.26 \\
        & MAXCUT    & 0.98          & \textbf{0.98} & \textbf{5.30} & \textbf{0.99} & 0.31 & 49.34 & 0.71 & 0.38 & 24.69 & 0.17 & 0.09 & 86.89 \\
        & MAXCLIQUE & \textbf{0.98} & \textbf{0.90} & \textbf{4.06} & 0.89 & 0.74 & 48.69 & 0.63 & 0.55 & 24.64 & 0.75 & 0.39 & 87.56 \\
        \bottomrule
    \end{tabular}%

    \caption{Benchmarking OptiDSL and baselines on the OptiDSLBench dataset, comprising 44 COP types and 4400 instances. Each cell presents the Execution Rate (\textit{ER}), Optimality Rate (\textit{OR}), and Modeling Time (\textit{MT}) averaged across the 100 test instances for each problem type. The best results are highlighted in \textbf{bold}.}
    \label{tab:exp1}
\end{table*}

\section{Experiments}

\begin{figure*}[t]
    \centering
    \begin{subfigure}[t]{0.35\linewidth}
        \centering
        \includegraphics[width=\linewidth]{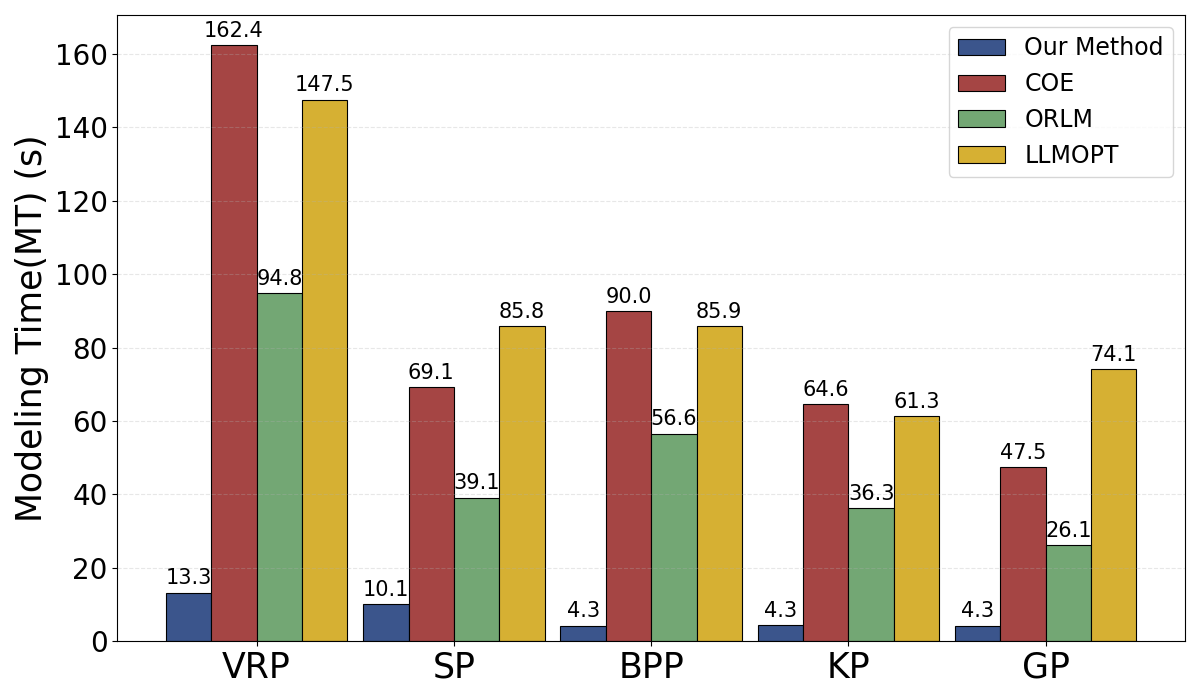}
        \caption{Modeling Time}
        \label{fig:exp1}
    \end{subfigure}
    \hspace{0.1\linewidth}
    \begin{subfigure}[t]{0.35\linewidth}
        \centering
        \includegraphics[width=\linewidth]{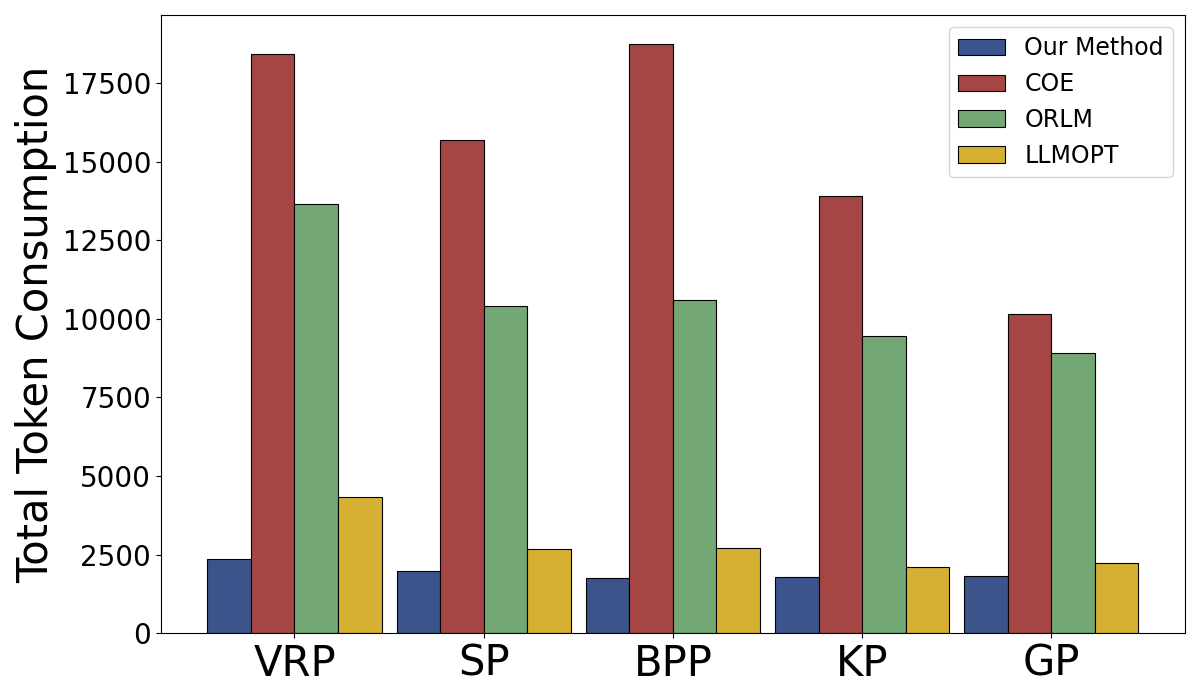}
        \caption{Total Token Consumption}
        \label{fig:token_analysis}
    \end{subfigure}
    \caption{Comparison of modeling time and token consumption across different domains.}
    \label{fig:efficiency_analysis}
\end{figure*}

\begin{figure}[t]
  \centering
  \includegraphics[width=0.7\linewidth]{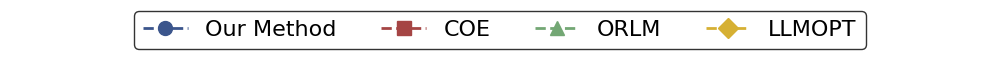} \\
  \begin{subfigure}{0.45\linewidth} 
     \includegraphics[width=\linewidth]{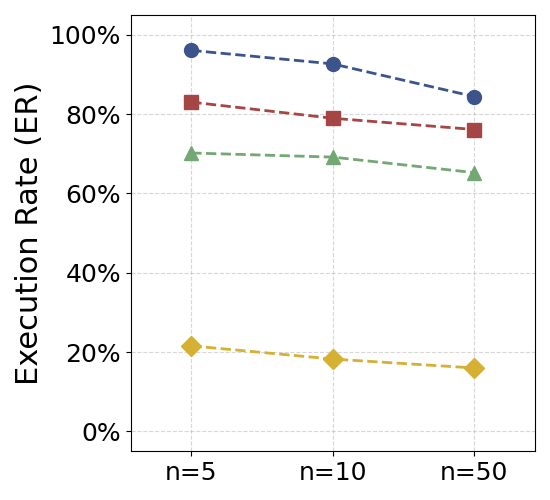}
     \caption{Impact of $n$ on ER}
  \end{subfigure}
  \begin{subfigure}{0.45\linewidth}
     \includegraphics[width=\linewidth]{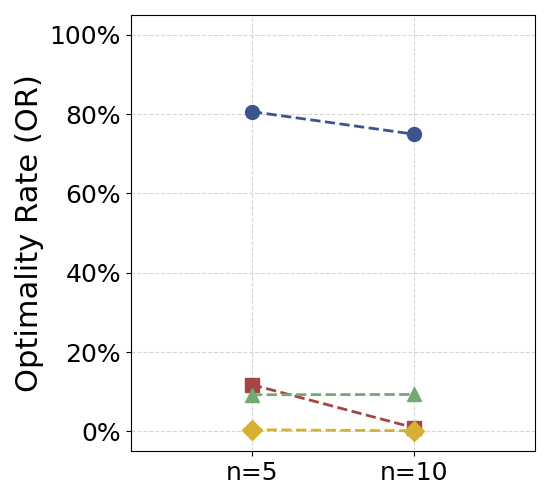}
     \caption{Impact of $n$ on OR}
  \end{subfigure}
  \caption{Performance of different methods on CVRP across varying node counts $n$.}
  \label{fig:sca}
\end{figure}

\subsection{Experimental Setup} 
We evaluate OptiDSL on our proposed \textit{OptiDSLBench} (44 COP types, 100 instances each) against three representative baselines: Chain-of-Experts~\citep{CoE}, ORLM~\citep{ORLM}, and LLMOPT~\citep{jiang2025llmopt}. OptiDSL and CoE use DeepSeek-V3.2~\citep{deepseekr1}, while ORLM and LLMOPT rely on their specific fine-tuned Llama3 and Qwen2.5-14B models, respectively.

Following LLMOPT~\citep{jiang2025llmopt}, we assess modeling performance using \textit{Execution Rate} (ER) and \textit{Optimality Rate} (OR). For $N$ instances, $\text{ER} = \frac{1}{N} \sum_{i=1}^N \mathbb{I}(\text{exec}_i)$, where $\mathbb{I}(\text{exec}_i) = 1$ if the formulation is successfully parsed and executed, reflecting structural reliability. Similarly, $\text{OR} = \frac{1}{N} \sum_{i=1}^N \mathbb{I}(\text{opt}_i)$, where $\mathbb{I}(\text{opt}_i) = 1$ if the formulation yields the optimal solution. Because baseline methods exclusively support MILP formats, we adopt Gurobi as the unified downstream solver across all evaluations to eliminate solver-dependent bias and ensure a fair comparison of formulation quality.

\subsection{Overall Performance Comparison}\label{sec:main_exp}
To assess the effectiveness of OptiDSL, we compare it with three representative baselines: Chain-of-Experts (CoE), ORLM, and LLMOPT. The evaluation is conducted on \textit{OptiDSLBench}, a benchmark introduced in this work that covers five problem domains. To ensure a fair comparison at the modeling stage, we \textbf{use Gurobi as the unified downstream solver} for all problems. Furthermore, to ensure that all evaluation instances can be solved to optimality, we set the problem size to 5 for the VRP and 10 for all other problems. The experimental results are summarized in Table~\ref{tab:exp1}.

\textbf{Superior Executability and Optimality.} 
Our method (OptiDSL) consistently maintains high ER and OR across all problem domains. Experimental results on OptiDSLBench demonstrate that, compared to baseline methods, OptiDSL achieves a 10.13\% improvement in average ER and a significant 51.66\% increase in average OR. Specifically, our approach shows the most substantial gains in the VRP domain, with ER and OR increasing by 13.05\% and 68.83\%, respectively. This clearly underscores the superiority of our method in complex problem domains like VRP, which are traditionally difficult to formulate as a standard five-elements.

\begin{table}[t] 
    \centering
    \renewcommand{\arraystretch}{0.88}
    \small
    \setlength{\tabcolsep}{2.5pt}
    \begin{tabular}{lcccccccc}
        \toprule
        \multirow{2}{*}{Method} 
        & \multicolumn{2}{c}{CVRP} 
        & \multicolumn{2}{c}{JSSP} 
        & \multicolumn{2}{c}{MIS} 
        & \multicolumn{2}{c}{MVC} \\
        \cmidrule(lr){2-3} \cmidrule(lr){4-5} \cmidrule(lr){6-7} \cmidrule(lr){8-9}
        & ER$\uparrow$ & OR$\uparrow$ 
        & ER$\uparrow$ & OR$\uparrow$ 
        & ER$\uparrow$ & OR$\uparrow$ 
        & ER$\uparrow$ & OR$\uparrow$ \\ 
        \midrule
        COE     & 0.87          & 0.03          & 0.42          & 0.14          & \textbf{0.96} & 0.88          & 0.91          & 0.85 \\
        ORLM    & 0.72          & 0.05          & 0.31          & 0.04          & 0.79          & 0.68          & 0.83          & 0.63 \\
        LLMOPT  & 0.06          & 0             & 0.56          & 0             & 0.81          & 0.47          & 0.77          & 0.32 \\
        \midrule 
        OptiDSL(Ours) & \textbf{0.94} & \textbf{0.83} & \textbf{1.00}    & \textbf{0.87} & \textbf{0.96} & \textbf{0.96} & \textbf{0.92} & \textbf{0.90} \\
        \bottomrule
    \end{tabular}
    \caption{Performance comparison on existing COP benchmark~\citep{jiang2025large}.}
    \label{tab:exp_cop}
\end{table}
    
\textbf{Broad Generalization Across Problem Domains.} Modeling difficulty varies significantly across different problem domains, and baseline methods often suffer severe performance degradation in complex scenarios. This is exemplified by the VRP, where even the best-performing baseline yields an ER of only $83.04\%$ and a critically low OR of $11.79\%$. In contrast, our method demonstrates exceptional robustness: across five distinct problem domains, we consistently maintain an ER above $95\%$ and an OR exceeding $80\%$. This clearly demonstrates the high generalization ability of our approach.

\textbf{Significant Computational and Resource Efficiency.}
A defining advantage of our method is the drastic reduction in both Modeling Time and Token Consumption. While baseline methods achieve an average optimal modeling time of 69.52 s across all problem domains, our method completes modeling in only 9.89 s (see Figure~\ref{fig:exp1}). Furthermore, by utilizing a concise DSL datafile representation, our approach significantly reduces token usage—particularly output tokens—compared to verbose MILP formulations (As shown in Figure~\ref{fig:token_analysis}). This makes the framework highly cost-effective and suitable for large-scale, time-sensitive industrial applications.

\subsection{Scalability Analysis}
To evaluate the scalability of OptiDSL, we conducted experiments on the CVRP with increasing numbers of nodes, i.e., $n \in \{5, 10, 50\}$. As shown in Figure~\ref{fig:sca}, OR is not reported for $n=50$, since computing global optima at this scale is intractable. As the problem size increases, both ER and OR exhibit a decreasing trend, indicating that larger instances are more challenging. Nevertheless, OptiDSL consistently outperforms all baselines and maintains a clear advantage across scales. Specifically, OptiDSL achieves an ER above 84\% even at $n=50$, while the best baseline attains only about 76\%. Moreover, at $n=10$, OptiDSL still obtains an OR of approximately 75\%, compared with only 9\% for the strongest baseline. These results demonstrate the superior robustness and scalability of OptiDSL.

\subsection{Modeling Performance on Existing Benchmarks}
To further evaluate the modeling performance of OptiDSL, we conduct experiments on both existing COP and MILP benchmarks. For COP benchmarks, we use the dataset derived from LLMCoSolver~\citep{jiang2025large}, covering CVRP, JSSP, MIS, and MVC. Although this dataset was originally tailored for learning the direct mapping from natural language descriptions to solutions, we repurposed its diverse problem descriptions as a robust testbed to assess OptiDSL's modeling capabilities. As shown in Table~\ref{tab:exp_cop}, OptiDSL achieves the best average performance, with ER of 0.94 and OR of 0.89, outperforming the strongest baselines by 10.96\% in ER and 23.09\% in OR. The gains are particularly notable on CVRP and JSSP, indicating that OptiDSL produces more reliable and accurate formulations across diverse COP domains.

We also evaluate OptiDSL on existing MILP-oriented benchmarks, including NL4Opt, MamoComplex, and NLP4LP. Since these benchmarks mainly focus on MILP-style formulations, we first filter COP-related instances and use Gurobi as the unified downstream solver for all methods to ensure a fair modeling-stage comparison. As shown in Table~\ref{tab:exp_milp}, OptiDSL achieves the highest average OR of 92.3\%, outperforming CoE, ORLM, and LLMOPT by 10.2, 32.3, and 43.6 percentage points, respectively. These results further show that OptiDSL improves modeling accuracy over generic MILP-style formulation methods.




\begin{table}[t]
\centering
\small

\begin{tabular}{lccc}
\toprule
\textbf{Type}
& \textbf{NL4Opt}
& \textbf{MamoComplex}
& \textbf{NLP4LP} \\
\midrule

CoE
& 100.0 (6/6)
& 75.8 (22/29)
& 100.0 (4/4) \\

ORLM
& 100.0 (6/6)
& 44.8 (13/29)
& 100.0 (4/4) \\

LLMOPT
& 83.3 (5/6)
& 37.9 (11/29)
& 75.0 (3/4) \\

\textbf{OptiDSL}
& \textbf{100.0 (6/6)}
& \textbf{89.6 (26/29)}
& \textbf{100.0 (4/4)} \\

\bottomrule
\end{tabular}

\caption{Performance comparison on existing MILP benchmarks in terms of optimality rate (OR).}
\label{tab:exp_milp}
\end{table}

\subsection{Downstream Solver Analysis}

Unlike baselines with rigid dependencies on specific solvers, OptiDSL employs a solver-agnostic paradigm that decouples modeling from solving. Table~\ref{tab:solver_cvrp} highlights the necessity of this flexibility: while exact solvers (Gurobi) are optimal for small-scale instances with negligible latency, they become computationally intractable as complexity scales. In large-scale regimes ($n \ge 50$), OptiDSL enables a strategic trade-off based on deployment needs: users can prioritize the learning-based solvers (RouteFinder) for time-critical scenarios, or prioritize \textit{solution precision} using Heuristics (PyVRP) to ensure near-optimal performance. This architectural decoupling ensures that OptiDSL maintains tractability and high solution quality across varying problem scales, a capability absent in static code generation methods.

\begin{table}[htbp]
\centering
\small
\setlength{\tabcolsep}{3pt}
\begin{tabular}{lcccccc}
\toprule
\multirow{2}{*}{\textbf{Size}} 
& \multicolumn{2}{c}{\textbf{Gurobi}} 
& \multicolumn{2}{c}{\textbf{PyVRP}} 
& \multicolumn{2}{c}{\textbf{RouteFinder}} \\
\cmidrule(lr){2-3} \cmidrule(lr){4-5} \cmidrule(lr){6-7}
& \textbf{Obj. $\downarrow$} & \textbf{Time / s}
& \textbf{Obj. $\downarrow$} & \textbf{Time / s}
& \textbf{Obj. $\downarrow$} & \textbf{Time / s} \\
\midrule
$n=5$   & 2.44 & 0.22   & 2.45  & 7.56  & 2.45  & 0.05 \\
$n=10$  & 3.76 & 0.52   & 3.77  & 10.59 & 3.89  & 0.07 \\
$n=50$  & --   & $>200$ & 10.12 & 15.57 & 10.32 & 0.23 \\
$n=100$ & --   & $>200$ & 14.58 & 25.60 & 14.98 & 0.56 \\
\bottomrule
\end{tabular}

\caption{Solver comparison on CVRP with different problem sizes.}
\label{tab:solver_cvrp}
\end{table}
\section{Conclusion}\label{sec:conclusion}

In this work, we introduce OptiDSL, a comprehensive framework for COPs. Leveraging LLMs, we construct a diverse benchmark OptiDSLBench encompassing 44 distinct COP types across 5 problem domains. Distinct from traditional MILP-based baselines, OptiDSL models problems as DSL data files. Empirical experiments demonstrate that our modeling approach achieves significant advantages in both solution quality and computational efficiency. Furthermore, OptiDSL exhibits exceptional adaptability by seamlessly supporting a diverse array of domain-specific solvers. This flexibility enables users to navigate the trade-off between computational speed and solution quality, allowing for precise matching of solvers to varying practical requirements.



\bibliography{aaai2027}

@inproceedings{NL4Opt,
  title = 	 {NL4Opt Competition: Formulating Optimization Problems Based on Their Natural Language Descriptions},
  author =       {Ramamonjison, Rindranirina and Yu, Timothy and Li, Raymond and Li, Haley and Carenini, Giuseppe and Ghaddar, Bissan and He, Shiqi and Mostajabdaveh, Mahdi and Banitalebi-Dehkordi, Amin and Zhou, Zirui and Zhang, Yong},
  booktitle = 	 {Proceedings of the NeurIPS 2022 Competitions Track},
  pages = 	 {189--203},
  year = 	 {2022},
  volume = 	 {220},
  series = 	 {Proceedings of Machine Learning Research},
  publisher =    {PMLR},
}

@inproceedings{CoE,
title={Chain-of-Experts: When {LLM}s Meet Complex Operations Research Problems},
author={Ziyang Xiao and Dongxiang Zhang and Yangjun Wu and Lilin Xu and Yuan Jessica Wang and Xiongwei Han and Xiaojin Fu and Tao Zhong and Jia Zeng and Mingli Song and Gang Chen},
booktitle={The Twelfth International Conference on Learning Representations},
year={2024},
}

@InProceedings{OptiMUS,
  title = 	 {{O}pti{MUS}: Scalable Optimization Modeling with ({MI}){LP} Solvers and Large Language Models},
  author =       {Ahmaditeshnizi, Ali and Gao, Wenzhi and Udell, Madeleine},
  booktitle = 	 {Proceedings of the 41st International Conference on Machine Learning},
  pages = 	 {577--596},
  year = 	 {2024},
  volume = 	 {235},
  series = 	 {Proceedings of Machine Learning Research},
  publisher =    {PMLR},
}

@article{ORLM,
	title = {ORLM: A Customizable Framework in Training Large Models for Automated Optimization Modeling},
	issn = {0030-364X},
	journal = {Operations Research},
	author = {Huang, Chenyu and Tang, Zhengyang and Hu, Shixi and Jiang, Ruoqing and Zheng, Xin and Ge, Dongdong and Wang, Benyou and Wang, Zizhuo},
	year = {2025},
}

@inproceedings{MAMO,
    title = {LLMs for Mathematical Modeling: Towards Bridging the Gap between Natural and Mathematical Languages},
    author = {Huang, Xuhan  and
      Shen, Qingning  and
      Hu, Yan  and
      Gao, Anningzhe  and
      Wang, Benyou},
    booktitle = {Findings of the Association for Computational Linguistics: NAACL 2025},
    year = {2025},
    pages = {2678--2710},
    ISBN = {979-8-89176-195-7}
}

@article{sohrabi2023genetic,
  title={Genetic Engineering Algorithm (GEA): An Efficient Metaheuristic Algorithm for Solving Combinatorial Optimization Problems},
  author={Sohrabi, Majid and Fathollahi-Fard, Amir M. and Gromov, Vasilii A.},
  journal={Avtomatika i Telemekhanika},
  volume={85},
  number={3},
  pages={23-37},
  year={2024},
}

@article{mu2016Simulated,
  title={Solving Vehicle Routing Problem with Simultaneous Pickup and Delivery Using Parallel Simulated Annealing Algorithm},
  author={Mu, Dong and Wang, Chao and Zhao, Fu and Sutherland, John W},
  journal={International Journal of Shipping and Transport Logistics},
  volume={8},
  number={1},
  pages={81--106},
  year={2016},
  publisher={Inderscience Publishers},
}

@phdthesis{Flerova2015,
  author       = {Natalia Flerova},
  title        = {Methods for Advancing Combinatorial Optimization Over Graphical Models},
  school       = {University of California, Irvine},
  year         = {2015}
}

@article{Consolini2019,
  author    = {Luca Consolini and Mattia Laurini and Marco Locatelli},
  title     = {Graph-based Algorithms for the Efficient Solution of Optimization Problems Involving Monotone Functions},
  journal   = {Computational Optimization and Applications},
  volume    = {73},
  number    = {1},
  pages     = {101--128},
  year      = {2019}
}

@inproceedings{berto2024routefinder,
    title={{RouteFinder}: Towards Foundation Models for Vehicle Routing Problems},
    author={Berto, Federico and Hua, Chuanbo and Zepeda, Nayeli Gast and Hottung, Andr{\'e} and Wouda, Niels and Lan, Leon and Tierney, Kevin and Park, Jinkyoo},
    booktitle={ICML 2024 Workshop on Foundation Models in the Wild},
    year={2024},
}

@inproceedings{sun2023difusco,
  title={DIFUSCO: Graph-based Diffusion Solvers for Combinatorial Optimization},
  author={Zhiqing Sun and Yiming Yang},
  booktitle={Proceedings of the 37th Conference on Neural Information Processing Systems (NeurIPS 2023)},
  year={2023}
}

@article{Jin2024Unified,
  author = {Yaochu Jin and Xueming Yan and Shiqing Liu and Xiangyu Wang},
  title = {A Unified Framework for Combinatorial Optimization Based on Graph Neural Networks},
  journal = {arXiv preprint arXiv:2406.13125},
  year = {2024}
}

@inproceedings{bello2017neural,
  title={Neural combinatorial optimization with reinforcement learning},
  author={Bello, Irwan and Pham, Hieu and Le, Quoc V and Norouzi, Mohammad and Bengio, Samy},
  booktitle={Proceedings of the International Conference on Learning Representations (ICLR) Workshops},
  year={2017}
}

@inproceedings{kwon2020pomo,
  title={POMO: Policy Optimization with Multiple Optima for Reinforcement Learning},
  author={Kwon, Yeong-Dae and Choo, Jinho and Kim, Byoungjip and Yoon, Iljoo and Gwon, Youngjune and Min, Seungjai},
  booktitle={Advances in Neural Information Processing Systems},
  volume={33},
  pages={21188--21198},
  year={2020}
}

@article{pyvrp2023,
  year = {2024},
  volume = {36},
  number = {4},
  pages = {943-955},
  publisher = {INFORMS},
  author = {Niels A. Wouda and Leon Lan and Wouter Kool},
  title = {PyVRP: a high-performance VRP solver package},
  journal = {INFORMS Journal on Computing},
}

@misc{ortools2019,
  title={OR-Tools},
  author={{Google Optimization Tools}},
  howpublished={\url{https://developers.google.com/optimization}},
  year={2019}
}

@inproceedings{zhao2022learning,
title={Learning Efficient Online 3D Bin Packing on Packing Configuration Trees},
author={Hang Zhao and Kai Xu},
booktitle={International Conference on Learning Representations},
year={2022},
}

@article{deepseekr1,
  title={Deepseek-r1: Incentivizing reasoning capability in llms via reinforcement learning},
  author={Guo, Daya and Yang, Dejian and Zhang, Haowei and Song, Junxiao and Zhang, Ruoyu and Xu, Runxin and Zhu, Qihao and Ma, Shirong and Wang, Peiyi and Bi, Xiao and others},
  journal={arXiv preprint arXiv:2501.12948},
  year={2025}
}

@article{wang2021review,
  title={A review of reinforcement learning based intelligent optimization for manufacturing scheduling},
  author={Wang, Ling and Pan, Zixiao and Wang, Jingjing},
  journal={Complex System Modeling and Simulation},
  volume={1},
  number={4},
  pages={257--270},
  year={2021},
  publisher={TUP}
}

@article{zong2025DDS,
author = {Zong, Zefang and Feng, Tao and Wang, Jingwei and Xia, Tong and Li, Yong},
title = {Deep Reinforcement Learning for Demand-Driven Services in Logistics and Transportation Systems: A Survey},
year = {2025},
publisher = {Association for Computing Machinery},
volume = {19},
number = {4},
issn = {1556-4681},
journal = {ACM Trans. Knowl. Discov. Data},
articleno = {89},
numpages = {42},
}

@article{chowdhery2023palm,
  title={Palm: Scaling language modeling with pathways},
  author={Chowdhery, Aakanksha and Narang, Sharan and Devlin, Jacob and Bosma, Maarten and Mishra, Gaurav and Roberts, Adam and Barham, Paul and Chung, Hyung Won and Sutton, Charles and Gehrmann, Sebastian and others},
  journal={Journal of Machine Learning Research},
  volume={24},
  number={240},
  pages={1--113},
  year={2023}
}

@article{brown2020language,
  title={Language models are few-shot learners},
  author={Brown, Tom and Mann, Benjamin and Ryder, Nick and Subbiah, Melanie and Kaplan, Jared D and Dhariwal, Prafulla and Neelakantan, Arvind and Shyam, Pranav and Sastry, Girish and Askell, Amanda and others},
  journal={Advances in neural information processing systems},
  volume={33},
  pages={1877--1901},
  year={2020}
}

@manual{gurobi2024,
  author       = {Gurobi Optimization, LLC},
  title        = {Gurobi Optimizer Reference Manual},
  year         = {2024},
  url          = {https://www.gurobi.com}
}

@manual{cplex2009,
  author       = {IBM ILOG Cplex},
  title        = {User’s Manual for CPLEX},
  year         = {2009},
  edition      = {V12.1},
  publisher    = {International Business Machines Corporation},
  volume       = {46},
  number       = {53},
  pages        = {157}
}

@inproceedings{kool2018attention,
title={Attention, Learn to Solve Routing Problems!},
author={Wouter Kool and Herke van Hoof and Max Welling},
booktitle={International Conference on Learning Representations},
year={2019},
}

@article{dispatchingRules1982,
author = {John H. Blackstone and Don T. Phillips and Gary L. Hogg and},
title = {A state-of-the-art survey of dispatching rules for manufacturing job shop operations},
journal = {International Journal of Production Research},
volume = {20},
number = {1},
pages = {27--45},
year = {1982},
publisher = {Taylor \& Francis},
}

@inproceedings{L2D2020,
 author = {Zhang, Cong and Song, Wen and Cao, Zhiguang and Zhang, Jie and Tan, Puay Siew and Chi, Xu},
 booktitle = {Advances in Neural Information Processing Systems},
 pages = {1621--1632},
 publisher = {Curran Associates, Inc.},
 title = {Learning to Dispatch for Job Shop Scheduling via Deep Reinforcement Learning},
 volume = {33},
 year = {2020}
}

@ARTICLE{FJSPDRL2023,
  author={Song, Wen and Chen, Xinyang and Li, Qiqiang and Cao, Zhiguang},
  journal={IEEE Transactions on Industrial Informatics}, 
  title={Flexible Job-Shop Scheduling via Graph Neural Network and Deep Reinforcement Learning}, 
  year={2023},
  volume={19},
  number={2},
  pages={1600-1610},}

@ARTICLE{DANIEL2024,
  author={Wang, Runqing and Wang, Gang and Sun, Jian and Deng, Fang and Chen, Jie},
  journal={IEEE Transactions on Neural Networks and Learning Systems}, 
  title={Flexible Job Shop Scheduling via Dual Attention Network-Based Reinforcement Learning}, 
  year={2024},
  volume={35},
  number={3},
  pages={3091-3102},
}

@inproceedings{MatNet2021,
 author = {Kwon, Yeong-Dae and Choo, Jinho and Yoon, Iljoo and Park, Minah and Park, Duwon and Gwon, Youngjune},
 booktitle = {Advances in Neural Information Processing Systems},
 pages = {5138--5149},
 publisher = {Curran Associates, Inc.},
 title = {Matrix encoding networks for neural combinatorial optimization},
 volume = {34},
 year = {2021}
}

@inproceedings{goal2025,
title={{GOAL}: A Generalist Combinatorial Optimization Agent Learner},
author={Darko Drakulic and Sofia Michel and Jean-Marc Andreoli},
booktitle={The Thirteenth International Conference on Learning Representations},
year={2025},
}

@inproceedings{li2024fast,
title={Fast T2T: Optimization Consistency Speeds Up Diffusion-Based Training-to-Testing Solving for Combinatorial Optimization},
author={Yang Li and Jinpei Guo and Runzhong Wang and Hongyuan Zha and Junchi Yan},
booktitle={The Thirty-eighth Annual Conference on Neural Information Processing Systems},
year={2024},
}

@InProceedings{diffuco2024,
  title = 	 {A Diffusion Model Framework for Unsupervised Neural Combinatorial Optimization},
  author =       {Sanokowski, Sebastian and Hochreiter, Sepp and Lehner, Sebastian},
  booktitle = 	 {Proceedings of the 41st International Conference on Machine Learning},
  pages = 	 {43346--43367},
  year = 	 {2024},
  volume = 	 {235},
  series = 	 {Proceedings of Machine Learning Research},
  publisher =    {PMLR},
}

@book{lkh,
title = "An Extension of the Lin-Kernighan-Helsgaun TSP Solver for Constrained Traveling Salesman and Vehicle Routing Problems: Technical report",
author = {Keld Helsgaun},
year = {2017},
publisher={Roskilde Universitet},
}

@article{LLMHallucination2025,
author = {Huang, Lei and Yu, Weijiang and Ma, Weitao and Zhong, Weihong and Feng, Zhangyin and Wang, Haotian and Chen, Qianglong and Peng, Weihua and Feng, Xiaocheng and Qin, Bing and Liu, Ting},
title = {A Survey on Hallucination in Large Language Models: Principles, Taxonomy, Challenges, and Open Questions},
year = {2025},
issue_date = {March 2025},
publisher = {Association for Computing Machinery},
address = {New York, NY, USA},
volume = {43},
number = {2},
issn = {1046-8188},
journal = {ACM Trans. Inf. Syst.},
articleno = {42},
numpages = {55},
}

@article{DFJF1960,
author = {Miller, C. E. and Tucker, A. W. and Zemlin, R. A.},
title = {Integer Programming Formulation of Traveling Salesman Problems},
year = {1960},
issue_date = {Oct. 1960},
publisher = {Association for Computing Machinery},
volume = {7},
number = {4},
issn = {0004-5411},
journal = {Journal of the ACM},
pages = {326–329},
numpages = {4}
}

@inproceedings{jiang2025llmopt,
title={LLMOPT: Learning to Define and Solve General Optimization Problems from Scratch},
author={Caigao Jiang and Xiang Shu and Hong Qian and Xingyu Lu and Jun Zhou and Aimin Zhou and Yang Yu},
booktitle={The Thirteenth International Conference on Learning Representations},
year={2025},
}

@inproceedings{Perron2023cpsat,
  author       = {Laurent Perron and
                  Fr{\'{e}}d{\'{e}}ric Didier and
                  Steven Gay},
  title        = {The {CP-SAT-LP} Solver (Invited Talk)},
  booktitle    = {29th International Conference on Principles and Practice of Constraint
                  Programming, {CP} 2023, Toronto, Canada, August 27-31, 2023},
  series       = {LIPIcs},
  volume       = {280},
  pages        = {3:1--3:2},
  year         = {2023},
}

@article{Zhang2011spga,
title = {An effective genetic algorithm for the flexible job-shop scheduling problem},
journal = {Expert Systems with Applications},
volume = {38},
number = {4},
pages = {3563-3573},
year = {2011},
issn = {0957-4174},
author = {Guohui Zhang and Liang Gao and Yang Shi},
}

@article{HOPPER200134,
title = {An empirical investigation of meta-heuristic and heuristic algorithms for a 2D packing problem},
journal = {European Journal of Operational Research},
volume = {128},
number = {1},
pages = {34-57},
year = {2001},
issn = {0377-2217},
author = {E Hopper and B.C.H Turton},
}

@article{TAILLARD1993278,
title = {Benchmarks for basic scheduling problems},
journal = {European Journal of Operational Research},
volume = {64},
number = {2},
pages = {278-285},
year = {1993},
note = {Project Management anf Scheduling},
issn = {0377-2217},
author = {E. Taillard}
}

@article{ChuBeasley1998,
  author  = {Chu, P. C. and Beasley, J. E.},
  title   = {A Genetic Algorithm for the Multidimensional Knapsack Problem},
  journal = {Journal of Heuristics},
  volume  = {4},
  number  = {1},
  pages   = {63--86},
  year    = {1998},
}

@inproceedings{nr-aaai15,
      title = {The Network Data Repository with Interactive Graph Analytics and Visualization},
      author={Ryan A. Rossi and Nesreen K. Ahmed},
      booktitle = {Proceedings of the Twenty-Ninth AAAI Conference on Artificial Intelligence},
      url={http://networkrepository.com},
      year={2015}
  }

@inproceedings{
jiang2025large,
title={Large Language Models as End-to-end Combinatorial Optimization Solvers},
author={Xia Jiang and Yaoxin Wu and Minshuo Li and Zhiguang Cao and Yingqian Zhang},
booktitle={The Thirty-ninth Annual Conference on Neural Information Processing Systems},
year={2025},
url={https://openreview.net/forum?id=qr5uMEs6iR}
}


\IfFileExists{supplement.pdf}{
    \clearpage
    

\section*{Supplementary material}

# Supplementary Material for
# "Not All Problems Are Best Modeled as MILP: A DSL-Centric Framework for Flexible and Accurate Optimization Modeling"

## Overview of Solvers

In this section, we provide a detailed information of the solvers used for solving various COPs in our experiments. These solvers can be categorized into two groups as follows,

**Non-Learning solvers**: **LKH** (Helsgaun 2000) is a powerful solver for the TSP based on the Lin-Kernighan heuristic, which efficiently finds near-optimal solutions through flexible edge exchange strategies and local search, and is widely applied to large-scale TSPs and related combinatorial optimization problems. **PyVRP** (Wouda, Lan, and Kool 2024) is an open-source Python library specifically designed for solving VRPs, offering flexible modeling capabilities and efficient heuristics to support various VRP variants, including capacity constraints and time windows. **OR-Tools** (Google Optimization Tools 2019) is an open-source optimization toolkit developed by Google that supports solving a wide range of combinatorial optimization problems such as vehicle routing, scheduling, and constraint programming, featuring powerful solvers and multi-language APIs. **CP-SAT** (Perron, Didier, and Gay 2023) is a hybrid solver within OR-Tools that integrates constraint programming and Boolean satisfiability (SAT) techniques, designed to efficiently solve complex logical and combinatorial optimization problems, and is widely applied in scheduling, routing, and resource allocation domains. **Gurobi** (Gurobi Optimization 2024) is a high-performance and fast solver that supports linear programming, integer programming, and mixed-integer programming, utilizing advanced mathematical optimization techniques such as branch-and-bound, cutting planes, and heuristic algorithms to efficiently explore the solution space and find global optimal solutions. **Genetic Algorithm (GA)** (Yao 1999) is a stochastic optimization technique inspired by natural selection that iteratively improves solutions through selection, crossover, and mutation, often used with heuristics for complex problems like bin packing. **Dynamic Programming (DP)** solves optimization problems by breaking them into overlapping subproblems and reusing solutions based on optimal substructure. **MILP** formulates problems with linear objectives and constraints, including integer variables, enabling precise solutions in logistics and scheduling. **Dispatching Rules** (Blackstone, Phillips, and and 1982) are simple heuristic methods that quickly make priority-based decisions in dynamic scheduling and routing but do not guarantee optimality.

**Learning solvers**: **RouteFinder** (Berto et al. 2024) is a transformer-based multi-task reinforcement learning framework that efficiently models and generalizes across 48 VRP variants using mixed batch training, reward normalization, and adaptable fine-tuning. **PCT** (Zhao and Xu 2022) structures 3D bin packing actions into a tree and combines a Transformer with reinforcement learning to guide efficient and feasible packing, significantly improving generalization and performance. **L2D** (Zhang et al. 2020) is a reinforcement learning-based scheduling method that learns dispatching policies to handle task allocation in dynamic and complex environments. **FJSP-DRL** (Song et al. 2023) employs graph neural networks and deep reinforcement learning to solve flexible job shop scheduling problems by jointly considering task sequences and resource assignments. **DANIEL** (Wang et al. 2024) introduces a graph attention-based framework that models the structural relationships between jobs and machines to enhance policy generalization. **MatNet** (Kwon et al. 2021) focuses on job-machine matching using graph neural networks to capture scheduling dependencies and learn effective policies. **GOAL** (Drakulic, Michel, and Andreoli 2025) proposes a generalized order-agnostic learning framework that improves adaptability to different permutations, making it suitable for a wide range of scheduling optimization problems. **FastT2T** (Li et al. 2024) is an efficient Transformer-to-Transformer learning framework designed for unified combinatorial optimization tasks. It learns fast mappings from problem instances to high-quality solutions and demonstrates strong generalization and inference efficiency. **DiffUCO** (Sanokowski, Hochreiter, and Lehner 2024) is a diffusion-based approach to unified combinatorial optimization, which models the optimization process in continuous solution spaces and achieves effective approximations for complex combinatorial problems with strong empirical performance and generalization. **POMO** (Kwon et al. 2020) is a reinforcement learning framework that simultaneously explores multiple optimal solutions to improve performance and generalization in combinatorial optimization tasks.

## Dataset Statistics

To evaluate the effectiveness of DSL Modeling Workflow, we construct a comprehensive and high-quality dataset named OptiDSL. As illustrated in Table 1, we compare OptiDSL

Table 1: **Comparison of OptiDSL**: OptiDSL provides the largest sample size and the highest average number of problem parameters, reflecting greater instance complexity. It supports multi-dimensional parameter spaces for flexible scenario description and includes problem type recognition for precise modeling.

| Dataset | # Samples | # Parameters (avg.) | Multi-dimensional Parameters | Problem Recognition |
|---|---|---|---|---|
| NL4OPT | 1101 | 4.96 | × | × |
| NLP4LP | 269 | 7.64 | ✓ | × |
| IndustryOR | 100 | 20.23 | × | × |
| OR-Instruct-Data-3K | 3000 | 18.89 | × | × |
| MAMO | 1209 | 19.19 | × | × |
| ComplexOR | 18 | 15.72 | ✓ | × |
| **OptiDSL (ours)** | 4400 | 33.61 | ✓ | ✓ |

with several existing mainstream datasets in the Operations Research (OR) domain, including NL4OPT, NLP4LP, and IndustryOR.

- **Largest Sample Size**: With 4,400 samples, OptiDSL is nearly 1.5 times the size of OR-Instruct-Data-3K. This large-scale data ensures better generalization for model training and evaluation.
- **Higher Instance Complexity**: The average number of problem parameters in OptiDSL reaches 33.61. This indicates that our dataset contains more intricate constraints and decision variables, reflecting real-world complexity more accurately than previous benchmarks.

## Solver-Compatible Structure Extraction

Figure 1 illustrates the process of converting natural language descriptions into structured data files. The LLM maps data such as coordinates, requirements, and capacity from the description to standard fields; simultaneously, it parses business logic—for example, inferring parameters like OPEN_ROUTE: TRUE based on phrases like ”avoiding unnecessary return journeys to the warehouse.”

## Supplementary Data for Complete 24 VRP Types and Model Performance

We present the complete experimental results for 24 distinct Vehicle Routing Problem (VRP) variants in Table 2. Additionally, the performance of OptiDSL across various open-source and closed-source LLMs is presented in Table 3 and Table 4, respectively.

## Impact of Base LLM Backend.

Table 3 evaluates the modularity of OptiDSL across different base models, revealing a distinct trade-off between modeling efficiency and reasoning depth. Lightweight models like Llama3.3-70B prioritize speed, achieving the lowest Modeling Time suitable for latency-sensitive applications. Conversely, reasoning-enhanced models such as DeepSeek-R1 (660B) deliver state-of-the-art solution quality, attaining the highest Optimality Rates and Execution Rates at the cost of increased computational overhead. These results confirm that OptiDSL allows for flexible customization to fit specific computational budgets, effectively balancing inference latency with complex reasoning capabilities.

## Generative AI Usage Statement

LLMs were used both as experimental components and to assist with language editing. All AI-assisted content was reviewed and verified by the authors, who take full responsibility for the manuscript.

## Parameter Settings for Instance Generation

Now, we present the parameter settings used for generating instances of COPs in our experiments.

**CVRP**:We follow RouteFinder (Berto et al. 2024) for data generation. Locations of $m$ depots and $n$ customers are sampled uniformly from the $[0,1]^2$. The vehicle capacity $C$ is determined by the number of customers $n$, and is calculated as follows:

$$C = \begin{cases} 30 + \lfloor \frac{1000}{5} + \frac{n-1000}{33.3} \rfloor, & \text{if } 1000 < n \\ 30 + \lfloor \frac{n}{5} \rfloor, & \text{if } 20 < n \leq 1000 \\ 30. & \text{otherwise} \end{cases}$$

Customer demands are randomly sampled from the discrete set $\{1, 2, \dots, 9\}$. Then, the demands are divided by the vehicle capacity to scale them into the range $[0, 1]$. For each customer, the probabilities of having linehaul and backhaul demands are 0.8 and 0.2, respectively. For each customer $i \in \{0, \dots, n-1\}$, the time windows $[e_i, l_i]$ and service times $s_i$ are generated through the following procedure: service times $s_i$ are first sampled uniformly from $[0.15, 0.18]$, and time window lengths $t_i$ are drawn from $[0.18, 0.2]$. The maximum distance $d_{\max}$ from any depot $j \in \{0, \dots, m-1\}$ to customer $i$ is computed as $d_{\max} = \max_j(d_{ij})$. Using this, an upper bound $h_i = \frac{t_{\max} - s_i - t_i}{d_{\max} - 1}$ for the start time is determined. The start time $e_i$ is then calculated as $e_i = (1 + (h_i - 1) \cdot u_i) \cdot d_{\max}$, where $u_i \sim U(0,1)$ is a uniform random variable. Finally, the end time $l_i$ is given by $l_i = e_i + t_i$. We sample distance limit $L$ from $U(2 \cdot \max(d_{0i}), l_{\max})$, where $d_{0i}$ is the distance from the depot to customer $i$, and $l_{\max} = 3.0$ is a predefined upper bound. During the construction of the OptiDSLBench dataset, $n$ is set to 5.

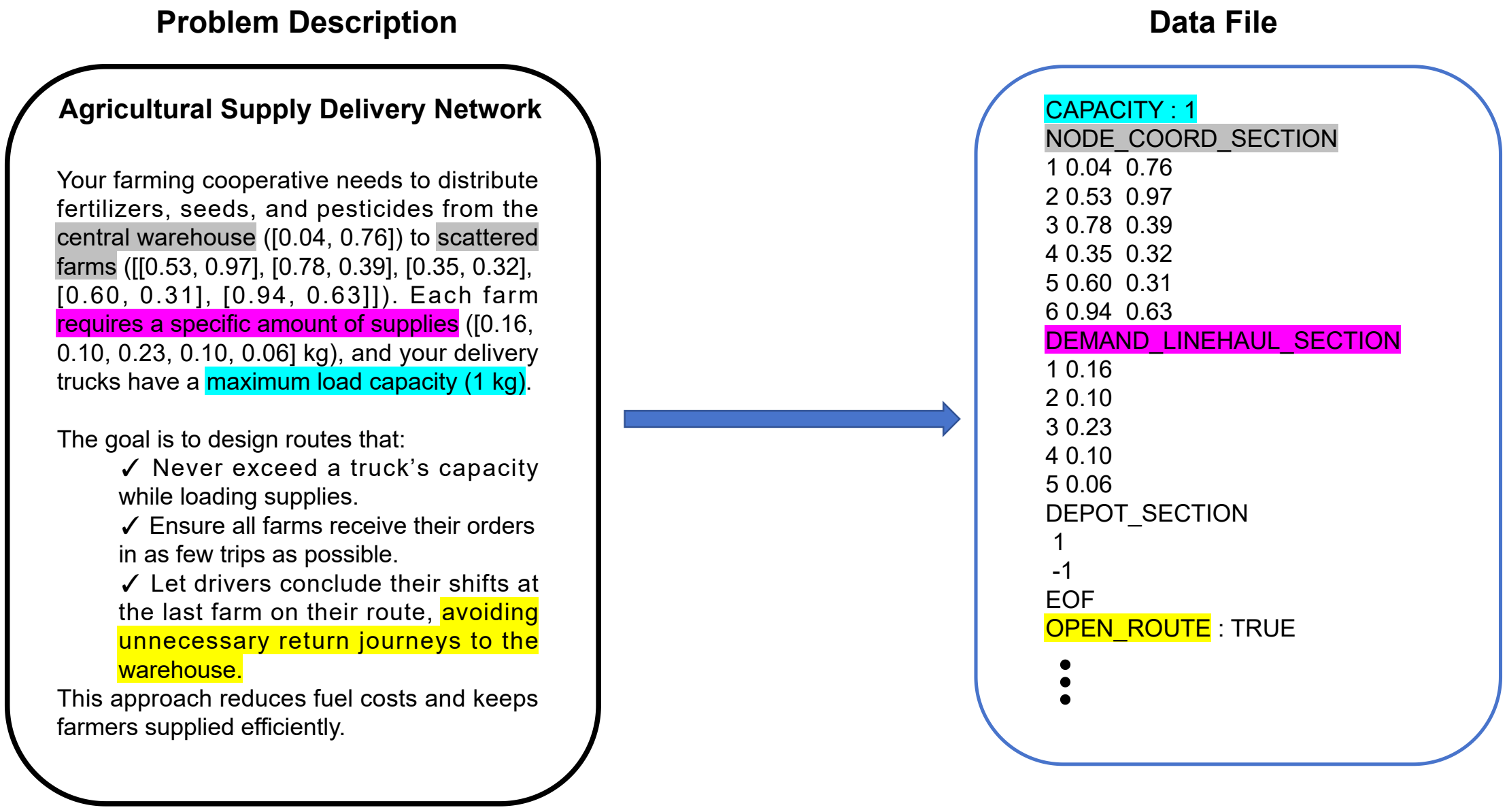


Figure 1: Transformation of natural language descriptions into structured data files via LLM

**Scheduling Problems (SP)**: For each SP instance, we randomly sample the number of jobs $J \in [4, 10]$ and machines $M \in [2, 5]$. For each job $j$, the number of operations $O_j$ is drawn from $[2, 5]$. The processing time of each operation $p_{j,k}$ is drawn from the integer range $[1, 10]$ (time units). All quantities are integers, and all ranges are inclusive.

**Bin Packing Problems (BPP)**: Each edge of the bins measures 10 in both the 2D and 3D cases. The size of each item is randomly generated, with each side length independently sampled from the discrete set $\{1, 2, \ldots, 10\}$. The number of items to be packed is set to 10.

**Graph Problems (GP)**: Following the DifFuCO (Sanokowski, Hochreiter, and Lehner 2024) setup, we generate MaxCut instances on Barabasi–Albert (BA) graphs (Barabási and Albert 1999), and MIS, MVC, and Max Clique instances on RB-model random graphs (Xu et al. 2005). The number of nodes is randomly sampled from the inclusive range $[9, 15]$.

**Knapsack Problems (KP)**: The experimental instances are generated with the following configuration. The number of items per instance is fixed at $N = 10$. The value $p_i$ and weight $w_i$ for each item $i$ are independently sampled from discrete uniform distributions, specifically $p_i \sim U(0, 20)$ and $w_i \sim U(0, 10)$. Regarding the constraints, the knapsack capacity $C$ is dynamically determined based on the total weight of the items, calculated as $C = \lfloor 0.6 \sum_{i=1}^{N} w_i \rfloor$. For multi-dimensional variants, an additional volume dimension $v_i \sim \mathcal{U}\{1, 5\}$ is introduced, with a corresponding volume capacity set to 60% of the total volume. The item availability limit $l_i$ is adapted to the problem type: $l_i = 1$ for 0-1 KP, $l_i \sim U(1, 5)$ for Bounded KP, and $l_i \to \infty$ for Unbounded KP.

## Implementation Details

All experiments are conducted on a single Ubuntu 22.04 LTS server equipped with an Intel Xeon Platinum 8378A CPU (3.00GHz), 8 NVIDIA A800-SXM4-40GB GPUs, and 1 TB of RAM.

## The Introduction of Benchmark Datasets

Below, we introduce several benchmark datasets that use natural language to describe optimization problems, particularly in the context of MILP.

**NL4OPT** (Ramamonjison et al. 2022) is a dataset developed for the NL4Opt competition, containing 1,101 annotated LPWPs from six distinct domains. The dataset is partitioned for training, development, and testing with 713, 99, and 289 samples, respectively. Instances from three source domains including sales, advertising, and investment are present across all splits, whereas instances from three target domains including production, transportation, and sciences are exclusively reserved for the development and test sets. This configuration facilitates a rigorous evaluation of model generalization and cross-domain transfer performance.

**NLP4LP** (Ahmaditeshnizi, Gao, and Udell 2024) contains problem instances sourced from textbooks and lecture notes in the field of optimization. It covers various types of optimization problems, including resource allocation, advertising placement, scheduling, and more. Each instance in the dataset includes a problem description, sample parameter data, and optimal value obtained either from textbook solutions or manual solving.

**IndustryOR** (Huang et al. 2025a) is the first industrial benchmark dataset consisting of 100 real-world operations

Table 2: Comparison of Different Methods on 24 VRP Types.

| Domain | Type | Our Method | | | COE | | | ORLM | | | LLMOPT | | |
|---|---|---|---|---|---|---|---|---|---|---|---|---|---|
| | | ER | OR↑ | MT↓ | ER | OR↑ | MT↓ | ER | OR↑ | MT↓ | ER | OR↑ | MT↓ |
| **VRP** | CVRP | **0.98** | **0.93** | **10.61** | 0.94 | 0.46 | 147.53 | 0.77 | 0.28 | 81.86 | 0.30 | 0.00 | 120.35 |
| | OVRP | **0.98** | **0.94** | **10.66** | 0.93 | 0.17 | 152.00 | 0.79 | 0.09 | 84.80 | 0.30 | 0.02 | 116.95 |
| | OVRPB | **0.91** | **0.60** | **11.47** | 0.86 | 0.04 | 144.20 | 0.76 | 0.03 | 90.33 | 0.28 | 0.00 | 128.40 |
| | OVRPBL | **0.87** | **0.51** | **11.70** | 0.85 | 0.02 | 240.53 | 0.75 | 0.01 | 155.02 | 0.22 | 0.00 | 151.96 |
| | VRPMBL | **0.86** | **0.68** | **11.52** | 0.81 | 0.18 | 166.08 | 0.70 | 0.14 | 94.62 | 0.18 | 0.00 | 130.23 |
| | VRPMBLTW | **0.98** | **0.88** | **15.59** | 0.78 | 0.06 | 160.38 | 0.66 | 0.08 | 99.95 | 0.12 | 0.00 | 171.91 |
| | VRPMBTW | **0.98** | **0.80** | **15.79** | 0.77 | 0.03 | 164.94 | 0.67 | 0.15 | 96.90 | 0.14 | 0.02 | 148.98 |
| | VRPTW | **1.00** | **0.99** | **15.45** | 0.82 | 0.15 | 155.18 | 0.61 | 0.02 | 91.21 | 0.10 | 0.00 | 161.23 |
| | OVRPBLTW | **0.97** | **0.74** | **16.42** | 0.67 | 0.04 | 189.70 | 0.58 | 0.03 | 104.23 | 0.24 | 0.00 | 155.99 |
| | OVRPBTW | **0.98** | **0.67** | **16.43** | 0.82 | 0.00 | 164.26 | 0.68 | 0.03 | 95.90 | 0.18 | 0.00 | 145.29 |
| | OVRPL | **0.98** | **0.96** | **10.56** | 0.96 | 0.17 | 148.38 | 0.82 | 0.14 | 87.48 | 0.26 | 0.02 | 138.72 |
| | OVRPLTW | **1.00** | **0.88** | **15.23** | 0.89 | 0.12 | 150.64 | 0.73 | 0.10 | 93.15 | 0.24 | 0.00 | 177.67 |
| | OVRPMB | **0.96** | **0.74** | **10.88** | 0.81 | 0.10 | 154.71 | 0.68 | 0.08 | 96.91 | 0.28 | 0.00 | 143.35 |
| | OVRPMBL | 0.78 | **0.55** | **11.29** | **0.84** | 0.09 | 151.26 | 0.74 | 0.07 | 79.39 | 0.36 | 0.00 | 140.49 |
| | OVRPMBLTW | **0.95** | **0.87** | **15.85** | 0.78 | 0.04 | 142.84 | 0.66 | 0.03 | 81.89 | 0.14 | 0.00 | 172.87 |
| | OVRPMBTW | **0.97** | **0.92** | **15.58** | 0.64 | 0.05 | 173.55 | 0.51 | 0.04 | 104.93 | 0.10 | 0.00 | 143.70 |
| | OVRPTW | **1.00** | **0.97** | **14.73** | 0.94 | 0.11 | 175.00 | 0.79 | 0.09 | 91.06 | 0.22 | 0.00 | 159.80 |
| | VRPB | **0.97** | **0.76** | **10.86** | 0.87 | 0.11 | 162.14 | 0.74 | 0.09 | 101.77 | 0.16 | 0.00 | 143.85 |
| | VRPBL | **0.98** | **0.73** | **11.01** | 0.91 | 0.07 | 165.70 | 0.75 | 0.05 | 95.78 | 0.20 | 0.00 | 141.30 |
| | VRPBLTW | **1.00** | **0.67** | **15.70** | 0.74 | 0.08 | 168.53 | 0.65 | 0.06 | 86.84 | 0.14 | 0.02 | 164.96 |
| | VRPBTW | **0.99** | **0.67** | **15.43** | 0.72 | 0.06 | 169.43 | 0.60 | 0.04 | 107.56 | 0.10 | 0.00 | 155.97 |
| | VRPL | **1.00** | **0.99** | **10.21** | 0.93 | 0.25 | 140.25 | 0.82 | 0.22 | 88.03 | 0.20 | 0.02 | 127.78 |
| | VRPLTW | **1.00** | **0.96** | **14.90** | 0.91 | 0.18 | 160.48 | 0.74 | 0.15 | 84.47 | 0.26 | 0.00 | 158.38 |
| | VRPMB | **0.97** | **0.94** | **10.70** | 0.74 | 0.25 | 150.93 | 0.65 | 0.20 | 82.02 | 0.46 | 0.00 | 140.64 |

Table 3: Comparison of Different Open-Source LLMs as Base Model on COPs.

| Model | Size | VRP | | | SP | | | BPP | | | GP | | | KP | | |
|---|---|---|---|---|---|---|---|---|---|---|---|---|---|---|---|---|
| | | ER | OR | MT | ER | OR | MT | ER | OR | MT | ER | OR | MT | ER | OR | MT |
| Llama3.3 | 70B | 0.96 | 0.40 | **7.17** | 0.45 | 0.40 | **4.09** | **1.00** | **0.97** | **2.92** | 0.87 | 0.61 | **2.58** | 0.94 | 0.53 | **3.09** |
| Qwen3 | 235B | 0.98 | 0.30 | 15.66 | 0.87 | 0.70 | 7.25 | **1.00** | 0.88 | 6.09 | **0.99** | **0.97** | 5.27 | 0.99 | **0.92** | 5.37 |
| DS-V3.2 | 671B | 0.96 | 0.81 | 13.29 | **1.00** | **0.85** | 6.77 | **1.00** | 0.95 | 4.29 | 0.98 | 0.96 | 4.28 | **1.00** | **0.92** | 4.34 |
| DS-R1 | 660B | **0.99** | **0.88** | 178.09 | 0.62 | 0.40 | 118.44 | **1.00** | **0.97** | 69.93 | 0.95 | 0.95 | 45.16 | 0.96 | 0.91 | 55.78 |

Table 4: Comparison of Different Closed-Source LLMs as Base Models on COPs.

| Model | CVRP | | | JSSP | | | 2DBPP | | | MDBKP | | | MAXCLIQUE | | |
|---|---|---|---|---|---|---|---|---|---|---|---|---|---|---|---|
| | ER | OR | MT | ER | OR | MT | ER | OR | MT | ER | OR | MT | ER | OR | MT |
| Claude-Sonnet-4.6 | 90 | 77 | 5.95 | **100** | 91 | 4.17 | **100** | **100** | 3.74 | **100** | **99** | 3.59 | 98 | 88 | 4.13 |
| Gemini-3-Flash | **99** | **85** | **4.29** | 98 | **97** | 3.25 | **100** | 99 | 3.76 | **100** | 98 | 3.22 | **100** | **97** | 2.95 |
| GPT-5.4 | 98 | 82 | 4.32 | 90 | 85 | **2.94** | **100** | **100** | **2.17** | **100** | 98 | **2.13** | **100** | 95 | **2.07** |

research problems collected from eight different industries. It covers five types of optimization problems including linear programming, integer programming, mixed integer programming, nonlinear programming, and others. The problems are categorized into three difficulty levels. Each instance includes a problem description, the optimal solution, and its difficulty level.

**OR-Instruct-Data-3K** (Huang et al. 2025a) is a sample consisting of 3000 instances selected from the OR-INSTRUCT dataset. OR-INSTRUCT dataset is a semi-automatically generated dataset produced by the OR-INSTRUCT framework and is commonly used as training data for LLMs in operations research and optimization.

**MAMO** (Huang et al. 2025b) is a benchmark dataset for mathematical modeling, comprising 346 ordinary differential equation (ODE) problems and 863 LP problems. The ODE problems are categorized into three types: 196 based on first-order differential equations, 110 based on second-

order differential equations, and 40 involving systems of differential equations. The LP problems are divided into two difficulty levels: the Easy LP subset contains 652 MILP problems designed for the high school level, while the Complex LP subset includes 211 undergraduate-level problems that integrate both LP and MILP models. Each problem instance is provided with a description and its optimal solution.

**ComplexOR** (Xiao et al. 2024) is the first dataset for complex operations research problems, constructed with the assistance of three experts in the field of operations research. The instances are sourced from academic papers, textbooks, and real-world industrial scenarios, covering multiple areas such as supply chain optimization, scheduling, and warehousing logistics.

## Examples of OptiDSL Problem Formulations

**VRPL**

**Problem Statement:**
A healthcare NGO must deliver critical medical supplies from the central warehouse located at [0.131, 0.977] to remote clinics situated at [[0.389, 0.560], [0.993, 0.860], [0.772, 0.004], [0.993, 0.644], [0.698, 0.244]] locations. Each clinic has a demand of [0.266, 0.266, 0.300, 0.033, 0.066] kg of medical equipment. The delivery vehicles have a maximum capacity of [1.0] kg. Due to fuel constraints and difficult terrain, each vehicle must complete its route without exceeding a round-trip distance limit of [2.424] km. Design vehicle routes to ensure all clinics receive their supplies while respecting vehicle capacity and distance constraints.

**DSL Formulation:**

```
TYPE: CVRP
DIMENSION: 6
EDGE_WEIGHT_TYPE: EUC_2D
CAPACITY: 1.0
DISTANCE_LIMITATIONS: 2.424
OPEN_ROUTE: FALSE
NODE_COORD_SECTION
 1 0.131 0.977
 2 0.389 0.560
 3 0.993 0.860
 4 0.772 0.004
 5 0.993 0.644
 6 0.698 0.244
DEMAND_LINEHAUL_SECTION
 1 0.266
 2 0.266
 3 0.300
 4 0.033
 5 0.066
DEMAND_BACKHAUL_SECTION: FALSE
DEPOT_SECTION
 1
 -1
EOF
MIXED_BACKHAUL: FALSE
TIME_WINDOWS: FALSE
SERVICE_TIME: FALSE
```

**OVRPLTW**

**Problem Statement:**
A regional distribution center ([0.155, 0.126]) needs to deliver medical supplies to 5 healthcare clinics ([[0.296, 0.920], [0.622, 0.4392], [0.565, 0.004], [0.457, 0.134], [0.093, 0.417]]). Each clinic requires a specific quantity of supplies ([0.100, 0.300, 0.2666, 0.1666, 0.066]), and deliveries must occur within strict time windows ([[0.0, 4.599], [3.079, 3.273], [1.566, 1.753], [1.308, 1.493], [2.669, 2.857], [1.726, 1.915]]) to ensure operational readiness. Delivery vehicles have a maximum load capacity ([1.0]) and each vehicle's route cannot exceed a maximum travel distance ([2.696]) to maintain efficiency. Each delivery requires a fixed service time ([0.0, 0.175, 0.156, 0.160, 0.176, 0.153]) for unloading. Drivers start at the distribution center, deliver supplies to clinics in optimal sequences, and end their routes at the final customer without returning. The goal is to ensure all supplies are delivered within their time windows while minimizing total travel distance and adhering to vehicle capacity and distance constraints.

**DSL Formulation:**

```
TYPE: CVRP
DIMENSION: 6
EDGE_WEIGHT_TYPE: EUC_2D
CAPACITY: 1.0
DISTANCE_LIMITATIONS: 2.696
OPEN_ROUTE: TRUE
NODE_COORD_SECTION
1 0.155 0.126
2 0.296 0.920
3 0.622 0.439
4 0.565 0.004
5 0.457 0.134
6 0.093 0.417
DEMAND_LINEHAUL_SECTION
1 0.0
2 0.100
3 0.300
4 0.266
5 0.166
6 0.066
DEMAND_BACKHAUL_SECTION: FALSE
DEPOT_SECTION
 1
 -1
EOF
MIXED_BACKHAUL: FALSE
TIME_WINDOWS
1 0.0 4.599
2 3.079 3.273
3 1.566 1.753
4 1.308 1.493
5 2.669 2.857
6 1.726 1.915
SERVICE_TIME
1 0.0
2 0.175
3 0.156
4 0.160
5 0.176
6 0.153
```

**UKP**

**Problem Statement:**
An investor aims to allocate capital 34 across various financial assets, where each asset requires an investment amount [10, 7, 10, 4, 6, 1, 6, 5, 1, 8] and offers an expected return [1, 16, 13, 1, 13, 17, 19, 2, 20, 5], with the possibility of purchasing fractional shares of the same asset multiple times to maximize total returns without surpassing the capital limit.

**DSL Formulation:**

```
item_number: 10
weight_limit: 34
volume_limit: false
item_value:
1
16
13
1
13
17
19
2
20
5
item_weight:
10
7
10
4
6
1
6
5
1
8
item_volume: false
item_limit: false
```

## MDBKP

**Problem Statement:**
A military unit must prepare supply packages for aerial delivery to forward bases, where each supply type (e.g., ammunition, rations) has a weight [9, 1, 8, 7, 8, 10, 9, 4, 9, 7], a volume [1, 1, 1, 2, 5, 5, 1, 5, 5, 2], a priority score [10, 3, 17, 1, 18, 17, 18, 11, 3, 13], and a maximum available quantity [3, 5, 5, 3, 3, 1, 4, 3, 5, 1]. Each cargo plane has a strict maximum payload 43 and cargo hold volume 16. The challenge is to select the optimal mix and quantity of supplies to load into each aircraft to maximize the total priority of delivered goods while adhering to both weight and volume constraints, ensuring critical mission needs are met under logistical limitations.

---

**DSL Formulation:**

```
item_number: 10
weight_limit: 43
volume_limit: 16
item_value:
10
3
17
1
18
17
18
11
3
13
item_weight:
9
1
8
7
8
10
9
4
9
7
item_volume:
1
1
1
2
5
5
1
5
5
2
item_limit:
3
5
5
3
3
1
4
3
5
1
```

## MAXCLIQUE

**Problem Statement:**
In blockchain analysis, the [(0, 1), (0, 2), (1, 2), (3, 4), (3, 5), (3, 8), (3, 7), (4, 5), (4, 8), (5, 6), (5, 7), (5, 8), (6, 7), (6, 8), (7, 8)] represents transactional relationships between wallet addresses, where 1 indicates a direct transfer between wallets and 0 means no recorded interaction. The task is to uncover the largest group of wallets that have all transacted with each other, potentially exposing coordinated trading rings or illicit financial networks that regulators may investigate.

---

**DSL Formulation:**

```
TYPE: maxclique
num_nodes: 9
num_edges: 15
0 1
0 2
1 2
3 4
3 5
3 8
3 7
4 5
4 8
5 6
5 7
5 8
6 7
6 8
7 8
```

**3DBPPR**

**Problem Statement:**
In the pharmaceutical industry, temperature-sensitive vaccines, vials, and medical supplies must be transported in refrigerated containers with strict [10, 10, 10] constraints. These fragile items come in varied [[8, 4, 4], [3, 3, 9], [9, 10, 9], [6, 3, 5], [1, 9, 8], [3, 3, 6], [1, 1, 5], [9, 10, 10], [8, 5, 6], [10, 7, 3]] forms from cylindrical vials to rectangular packaging and must be rotated strategically to maximize space while maintaining temperature stability. Using the fewest possible containers is critical to reduce costs and ensure timely delivery of life-saving medications.

**DSL Formulation:**

```
name: 3DBPPR
bin_size: 10 10 10
rotation: TRUE
no      Length   width    height
1       8        4        4
2       3        3        9
3       9        10       9
4       6        3        5
5       1        9        8
6       3        3        6
7       1        1        5
8       9        10       10
9       8        5        6
10      10       7        3
```

## Prompt for generating description

The following prompt is used to generate diverse real-world scenarios for the 2D Multiple Knapsack Problem. It is structured to ensure consistent use of placeholders and varied linguistic styles.

```
Please provide 5 real-world application
    scenarios of the 2-dimensional Multiple
    Knapsack Problem, described in a single
    paragraph.

[Variable Requirements]
Each item has a weight <item_weight>, a
    volume <item_volume>, a value <
    item_value>, and a maximum quantity <
    item_limit>. The knapsack has two
    capacity limits: maximum weight <
    knapsack_capacity> and maximum volume <
    knapsack_volume>.

[Task Constraints]
- The goal is to find the optimal subset of
    items and quantities to maximize total
    value without exceeding weight or volume
     limits.
- Descriptions must be complete (definition,
     attributes, limits, objective, and
    significance).
- Language style must be varied and natural
    across examples.
- Placeholders MUST be used exactly as
    defined.
- NO DUPLICATES: The industry titles of the
    5 new scenarios MUST NOT duplicate or be
     similar to the following: [Military
    Supply Airdrop, Spacecraft Payload
    Configuration,......].

[Few-shot Examples]
# Example 1: Logistics
##Warehouse Order Fulfillment## A logistics
    company needs to pack multiple customer
    orders into delivery trucks, where each
    product has a weight <item_weight>, a
    volume <item_volume>, a profit <
    item_value>, and a maximum available
    stock <item_limit>. Each truck has
    maximum weight <knapsack_capacity> and
    maximum volume <knapsack_volume>
    restrictions. The goal is to select
    which products and how many units of
    each to load onto each truck to maximize
     total profit while not exceeding the
    trucks' weight and volume capacities.

&&&

# Example 2: Healthcare
##Hospital Medical Supply Allocation## A
    hospital is distributing medical
    supplies to multiple departments, where
    each supply type has a weight <
    item_weight>, a volume <item_volume>, a
    criticality value <item_value>, and a
    maximum stock limit <item_limit>. Each
    department's storage has maximum weight
    <knapsack_capacity> and volume <
    knapsack_volume>. The objective is to
    assign the optimal quantity of each
    supply to maximize total medical utility
     while respecting storage constraints.

&&&

# Example 3: Retail
##Retail Store Shelf Allocation## A retail
    chain needs to allocate products across
    multiple store shelves, where each
    product has a weight <item_weight>, a
    volume <item_volume>, a sales value <
    item_value>, and a maximum stock limit <
    item_limit>. Each shelf has maximum
    weight <knapsack_capacity> and volume <
    knapsack_volume>. The goal is to
    determine the quantities of each product
     to place on each shelf to maximize
    expected sales.

[Output Format]
- Start each scenario with ##Industry Title
    ##.
- Separate examples with the symbol &&&.
- Output EVERYTHING in a single paragraph
```

```
without line breaks.
```

## Prompt for DSL formulation

```
You are an information extraction system
.

Task:
Select the SINGLE matching template and
fill it with data.

Rules:
    -Output ONLY the completed template.
    -Keep field names, order, and
structure EXACTLY as in the template.
    -Do NOT add or remove any fields.
    -If a field is missing in the
problem, fill in FALSE.
    -The output must begin and end with
###.
    -No explanations or extra text.
    -Do not include any parentheses and
explanations in the output content

-----------------------------------------------------


Problem:
An investor aims to allocate capital 34
across various financial assets, where
each asset requires an investment amount
 [10, 7, 10, 4, 6, 1, 6, 5, 1, 8] and
offers an expected return [1, 16, 13, 1,
 13, 17, 19, 2, 20, 5], with the
possibility of purchasing fractional
shares of the same asset multiple times
to maximize total returns without
surpassing the capital limit.

Templates:
-------------------------------------
[Templates ]
.
.
.
```

## References


Ahmaditeshnizi, A.; Gao, W.; and Udell, M. 2024. OptiMUS: Scalable Optimization Modeling with (MI)LP Solvers and Large Language Models. In *Proceedings of the 41st International Conference on Machine Learning*, volume 235 of *Proceedings of Machine Learning Research*, 577–596. PMLR.

Barabási, A.-L.; and Albert, R. 1999. Emergence of Scaling in Random Networks. *Science*, 286(5439): 509–512.

Berto, F.; Hua, C.; Zepeda, N. G.; Hottung, A.; Wouda, N.; Lan, L.; Tierney, K.; and Park, J. 2024. RouteFinder: Towards Foundation Models for Vehicle Routing Problems. In *ICML 2024 Workshop on Foundation Models in the Wild*.

Blackstone, J. H.; Phillips, D. T.; and and, G. L. H. 1982. A state-of-the-art survey of dispatching rules for manufacturing job shop operations. *International Journal of Production Research*, 20(1): 27–45.

Drakulic, D.; Michel, S.; and Andreoli, J.-M. 2025. GOAL: A Generalist Combinatorial Optimization Agent Learner. In *The Thirteenth International Conference on Learning Representations*.

Google Optimization Tools. 2019. OR-Tools. https://developers.google.com/optimization.

Gurobi Optimization, L. 2024. *Gurobi Optimizer Reference Manual*.

Helsgaun, K. 2000. An Effective Implementation of the Lin-Kernighan Traveling Salesman Heuristic. Technical Report RR 2000-12, Roskilde University.

Huang, C.; Tang, Z.; Hu, S.; Jiang, R.; Zheng, X.; Ge, D.; Wang, B.; and Wang, Z. 2025a. ORLM: A Customizable Framework in Training Large Models for Automated Optimization Modeling. *Operations Research*.

Huang, X.; Shen, Q.; Hu, Y.; Gao, A.; and Wang, B. 2025b. LLMs for Mathematical Modeling: Towards Bridging the Gap between Natural and Mathematical Languages. In *Findings of the Association for Computational Linguistics: NAACL 2025*, 2678–2710. ISBN 979-8-89176-195-7.

Kwon, Y.-D.; Choo, J.; Kim, B.; Yoon, I.; Gwon, Y.; and Min, S. 2020. POMO: Policy Optimization with Multiple Optima for Reinforcement Learning. In *Advances in Neural Information Processing Systems*, volume 33, 21188–21198.

Kwon, Y.-D.; Choo, J.; Yoon, I.; Park, M.; Park, D.; and Gwon, Y. 2021. Matrix encoding networks for neural combinatorial optimization. In *Advances in Neural Information Processing Systems*, volume 34, 5138–5149. Curran Associates, Inc.

Li, Y.; Guo, J.; Wang, R.; Zha, H.; and Yan, J. 2024. Fast T2T: Optimization Consistency Speeds Up Diffusion-Based Training-to-Testing Solving for Combinatorial Optimization. In *The Thirty-eighth Annual Conference on Neural Information Processing Systems*.

Perron, L.; Didier, F.; and Gay, S. 2023. The CP-SAT-LP Solver (Invited Talk). In *29th International Conference on Principles and Practice of Constraint Programming, CP 2023, Toronto, Canada, August 27-31, 2023*, volume 280 of *LIPIcs*, 3:1–3:2.

Ramamonjison, R.; Yu, T.; Li, R.; Li, H.; Carenini, G.; Ghaddar, B.; He, S.; Mostajabdaveh, M.; Banitalebi-Dehkordi, A.; Zhou, Z.; and Zhang, Y. 2022. NL4Opt Competition: Formulating Optimization Problems Based on Their Natural Language Descriptions. In *Proceedings of the NeurIPS 2022 Competitions Track*, volume 220 of *Proceedings of Machine Learning Research*, 189–203. PMLR.

Sanokowski, S.; Hochreiter, S.; and Lehner, S. 2024. A Diffusion Model Framework for Unsupervised Neural Combinatorial Optimization. In *Proceedings of the 41st International Conference on Machine Learning*, volume 235 of *Proceedings of Machine Learning Research*, 43346–43367. PMLR.

Song, W.; Chen, X.; Li, Q.; and Cao, Z. 2023. Flexible Job-Shop Scheduling via Graph Neural Network and Deep Reinforcement Learning. *IEEE Transactions on Industrial Informatics*, 19(2): 1600–1610.

Wang, R.; Wang, G.; Sun, J.; Deng, F.; and Chen, J. 2024. Flexible Job Shop Scheduling via Dual Attention Network-Based Reinforcement Learning. *IEEE Transactions on Neural Networks and Learning Systems*, 35(3): 3091–3102.

Wouda, N. A.; Lan, L.; and Kool, W. 2024. PyVRP: a high-performance VRP solver package. *INFORMS Journal on Computing*, 36(4): 943–955.

Xiao, Z.; Zhang, D.; Wu, Y.; Xu, L.; Wang, Y. J.; Han, X.; Fu, X.; Zhong, T.; Zeng, J.; Song, M.; and Chen, G. 2024. Chain-of-Experts: When LLMs Meet Complex Operations Research Problems. In *The Twelfth International Conference on Learning Representations*.

Xu, K.; Boussemart, F.; Hemery, F.; and Lecoutre, C. 2005. A simple model to generate hard satisfiable instances. In *Proceedings of the 19th International Joint Conference on Artificial Intelligence*, IJCAI'05, 337–342. Morgan Kaufmann Publishers Inc.

Yao, X. 1999. *Evolutionary computation: Theory and applications*. World scientific.

Zhang, C.; Song, W.; Cao, Z.; Zhang, J.; Tan, P. S.; and Chi, X. 2020. Learning to Dispatch for Job Shop Scheduling via Deep Reinforcement Learning. In *Advances in Neural Information Processing Systems*, volume 33, 1621–1632. Curran Associates, Inc.

Zhao, H.; and Xu, K. 2022. Learning Efficient Online 3D Bin Packing on Packing Configuration Trees. In *International Conference on Learning Representations*.


}{
    \typeout{WARNING: supplement.pdf not found. Compile supplement.tex first to append it to aaai2027.pdf.}
}
\end{document}